\RequirePackage{fix-cm}
\documentclass[twocolumn]{svjour3}          
\smartqed  
\usepackage{graphicx}
\usepackage{algorithmic,algorithm}
\newcommand{\Algo}[1]{Algorithm~\ref{algo:#1}}

\newcommand{\algolabel}[1]{\label{algo:#1}}

\usepackage{hyperref}
\usepackage{amssymb}
\usepackage{amsmath}
\usepackage{color}
\usepackage[normalem]{ulem}
\usepackage{bm}
\usepackage{booktabs}
\usepackage{float}

\usepackage[round]{natbib}

\begin{document}

\title{Ensemble Slice Sampling
}
\subtitle{Parallel, black-box and gradient-free inference for correlated \& multimodal distributions}


\author{Minas Karamanis \and
        Florian Beutler 
}


\institute{M. Karamanis \at
              \email{minas.karamanis@ed.ac.uk}             \\
              Institute for Astronomy, University of Edinburgh, Royal Observatory, Blackford Hill, Edinburgh EH9 3HJ, UK 
           \and
           F. Beutler \at
           \email{florian.beutler@ed.ac.uk}\\
              Institute for Astronomy, University of Edinburgh, Royal Observatory, Blackford Hill, Edinburgh EH9 3HJ, UK
}

\date{Received: date / Accepted: date}

\maketitle

\begin{abstract}
Slice Sampling has emerged as a powerful Markov Chain Monte Carlo algorithm that adapts to the characteristics of the target distribution with minimal hand-tuning. However, Slice Sampling's performance is highly sensitive to the user-specified initial length scale hyperparameter and the method generally struggles with poorly scaled or strongly correlated distributions. This paper introduces Ensemble Slice Sampling (ESS), a new class of algorithms that bypasses such difficulties by adaptively tuning the initial length scale and utilising an ensemble of parallel walkers in order to efficiently handle strong correlations between parameters. These affine-invariant algorithms are trivial to construct, require no hand-tuning, and can easily be implemented in parallel computing environments. Empirical tests show that Ensemble Slice Sampling can improve efficiency by more than an order of magnitude compared to conventional MCMC methods on a broad range of highly correlated target distributions. In cases of strongly multimodal target distributions, Ensemble Slice Sampling can sample efficiently even in high dimensions. We argue that the parallel, black-box and gradient-free nature of the method renders it ideal for use in scientific fields such as physics, astrophysics and cosmology which are dominated by a wide variety of computationally expensive and non-differentiable models.
\keywords{Markov Chain Monte Carlo \and Bayesian Inference \and Slice Sampling \and Adaptive Monte Carlo \and Probabilistic Data Analysis}
\end{abstract}

\section{Introduction}
\emph{Bayesian inference and data analysis} has become an integral part of modern science. This is partly due to the ability of Markov Chain Monte Carlo (MCMC) algorithms to generate samples from intractable probability distributions. MCMC methods produce a sequence of samples, called a \emph{Markov chain}, that has the target distribution as its equilibrium distribution. The more samples are included, the more closely the distribution of the samples approaches the target distribution. The Markov chain can then be used to numerically approximate expectation values (e.g. parameter uncertainties, marginalised distributions).

Common MCMC methods entail a significant amount of time spent hand-tuning the hyperparameters of the algorithm to optimize its efficiency with respect to a target distribution. The emerging and routine use of such mathematical tools in science calls for the development of black-box MCMC algorithms that require no hand-tuning at all. This need led to the development of adaptive MCMC methods like the Adaptive Metropolis algorithm \citep{haario2001adaptive} which tunes its proposal scale based on the sample covariance matrix. Unfortunately, most of those algorithms still include a significant number of hyperparameters (e.g. components of the covariance matrix)  rendering the adaptation noisy. Furthermore, the tuning is usually performed on the basis of prior knowledge, such as one or more long preliminary runs which further slow down the sampling. Last but not least, there is no reason to believe that a single Metropolis proposal scale is optimal for the whole distribution (i.e. the appropriate scale could vary from one part of the distribution to another). Another approach to deal with those issues would be to develop methods that by construction require no or minimal hand-tuning. An archetypal such method is the Slice Sampler \citep{neal2003slice}, which has only one hyperparameter, the initial length scale.

It should be noted that powerful adaptive methods that require no hand-tuning (although they do require preliminary runs) already exist. Most notable of them is the No U-Turn Sampler (NUTS) \citep{hoffman2014no}, an adaptive extension of Hamiltonian Monte Carlo (HMC) \citep{neal2011mcmc}. However, such methods rely on the gradient of the log probability density function. This requirement is the reason why these methods are limited in their application in quantitative fields such as physics, astrophysics and cosmology, which are dominated by computationally expensive non-differentiable models. Thus, our objective in this paper is to introduce a parallel, black-box and gradient-free method that can be used in the aforementioned scientific fields.

This paper presents Ensemble Slice Sampling (ESS), an extension of the Standard Slice Sampling method. ESS naturally inherits most of the benefits of Standard Slice Sampling, such as the acceptance rate of $1$, and most importantly the ability to adapt to the characteristics of a target distribution without any hand-tuning at all. Furthermore, we will show that ESS's performance is insensitive to linear correlations between the parameters, thus enabling efficient sampling even in highly demanding scenarios. We will also demonstrate ESS's performance in strongly multimodal target distributions and show that the method samples efficiently even in high dimensions. Finally, the method can easily be implemented in parallel taking advantage of multiple CPUs thus facilitating Bayesian inference in cases of computationally expensive models.

Our implementation of ESS is inspired by \citet{tran2015reunderstanding}. However, our method improves upon that by extending the direction choices (e.g. Gaussian and global move), adaptively tuning the initial proposal scale, and parallelising the algorithm. \citet{nishihara2014parallel} developed a general algorithm based on the elliptical slice sampling method~\citep{murray2010elliptical} and a Gaussian mixture approximation to the target distribution. ESS utilises an ensemble of parallel and interacting chains, called walkers. Other methods that are based on the ensemble paradigm include the Affine Invariant Ensemble Sampler \citep{goodman2010ensemble} and the Differential Evolution MCMC \citep{ter2006markov} along with its various extensions \citep{ter2008differential, vrugt2009accelerating}, as well as more recent approaches that are based on langevin diffusion dynamics \citep{garbuno2020interacting, garbuno2020affine} and the time discretization of stochastic differential equations \citep{leimkuhler2018ensemble} in order to achieve substantial speedups.

In Section \ref{sec:slice}, we will briefly discuss the Standard Slice Sampling algorithm. In Section \ref{sec:ensemble}, we will introduce the Ensemble Slice Sampling method. In Section \ref{sec:empirical} we will investigate the empirical evaluation of the algorithm. We reserve Sections \ref{sec:discussion} and \ref{sec:conclusion} for discussion and conclusion, respectively.

\section{Standard Slice Sampling}
\label{sec:slice}
\textit{Slice Sampling} is based on the idea that sampling from a distribution $p(x)$ whose density is proportional to $f(x)$ is equivalent to uniformly sampling from the region underneath the graph of $f(x)$. More formally, in the univariate case, we introduce an auxiliary variable, the height $y$, thus defining the joint distribution $p(x,y)$, which is uniform over the region $U = \{ (x,y) : 0 < y < f(x) \}$. To sample from the marginal density for $x$, $p(x)$, we sample from $p(x,y)$ and then we ignore the $y$ values.

Generating samples from $p(x,y)$ is not trivial, so we might consider defining a Markov chain that will converge to that distribution. The simplest, in principle, way to construct such a Markov chain is via Gibbs sampling. Given the current $x$, we sample $y$ from the conditional distribution of $y$ given $x$, which is uniform over the range $(0, f(x) )$. Then we sample the new $x$ from the \textit{slice} $S=\{ x : y < f(x)\}$. 

Generating a sample from the slice $S$ may still be difficult, since we generally do not know the exact form of $S$. In that case, we can update $x$ based on a procedure that leaves the uniform distribution of $S$ invariant. \citet{neal2003slice} proposed the following method:
\begin{description}
\item Given the current state $x_{0}$, the next one is generated as:
\begin{enumerate}
    \item Draw $y_{0}$ uniformly from $(0,f(x_{0}))$, thus defining the horizontal slice $S = \{ x : y_{0} < f(x)\}$,
    \item Find an interval $I = (L,R)$ that contains all, or much, of $S$ (e.g. using the stepping-out procedure defined bellow),
    \item Draw the new point $x_{1}$ uniformly from $I \cap S$.
\end{enumerate}
\end{description}

\begin{figure}
    \centering
    \includegraphics[scale=0.45]{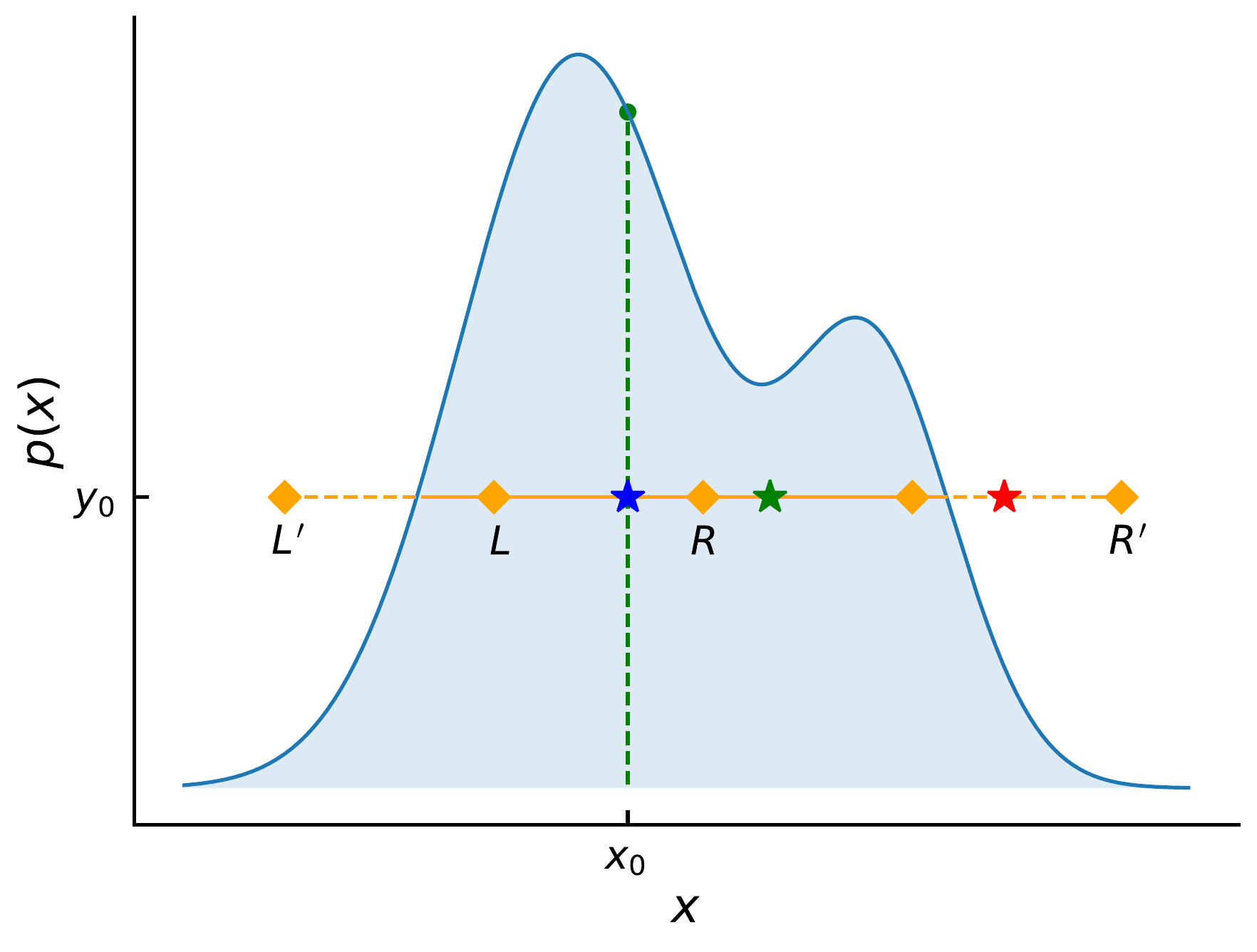}
    \caption{The plot shows the univariate slice sampling method. Given an initial value $x_{0}$, a value $y_{0}$ is uniformly sampled along the vertical slice $(0,f(x_{0}))$ (green dashed line) thus defining the initial point (blue star). An interval $(L,R)$ is randomly positioned horizontally around the initial point, and then it is expanded in steps of size $\mu=R-L$ until both of its ends $L', R'$ are outside the slice. The new point (green star) is generated by repeatedly sampling uniformly from the expanded interval $(L',R')$ until a point is found inside the slice. Points outside the slice (e.g. the red star) are used to shrink the interval $(L',R')$ by moving $L'$ or in this case $R'$ to that point and accelerate the sampling procedure.}
\label{fig:slice}
\end{figure}

In order to find the interval $I$, \citet{neal2003slice} proposed to use the \emph{stepping-out} procedure that works by randomly positioning an interval of length $\mu$ around the point $x_{0}$ and then expanding it in steps of size $\mu$ until both ends are outside of the slice. The new point $x_{1}$ is found using the \emph{shrinking} procedure, in which points are uniformly sampled from $I$ until a point inside $S$ is found. Points outside $S$ are used to shrink the interval $I$. The stepping-out and shrinking  procedures are illustrated in Figure \ref{fig:slice}. By construction, the stepping-out and shrinking procedures can adaptively tune a poor estimate of the length scale $\mu$ of the initial interval. The length scale $\mu$ is the only free hyperparameter of the algorithm. For a detailed review of the method we direct the reader to \cite{neal2003slice} and \cite{mackay2003information} (also Exercise 30.12 in that text).

It is important to mention here that for multimodal distributions there is no guarantee that the slice would cross any of the other modes, especially if the length scale is underestimated initially. Ideally, in order to provide a large enough initial value of the scale factor $\mu$, prior knowledge of the distance between the modes is required. As we will show in the next section, Ensemble Slice Sampling does not suffer from this complication and can handle strongly multimodal distributions efficiently.

\section{Ensemble Slice Sampling}
\label{sec:ensemble}
The univariate slice sampling scheme can be used to sample from multivariate distributions by sampling repeatedly along each coordinate axis in turn (one parameter at a time) or by sampling along randomly selected directions \citep{mackay2003information}. Using either of those choices, the Standard Slice Sampler performs acceptably in cases with no strong correlations in parameter space. The overall performance of the algorithm generally depends on the number of expansions and contractions during the stepping-out and shrinking procedures, respectively. Ideally we would like to minimize that number. A reasonable initial estimate of the length scale is still required in order to reduce the amount of time spent expanding or contracting the initial interval. 

However, when strong correlations are present two issues arise. First, there is no single value of the initial length scale that minimizes the computational cost of the stepping-out and shrinking procedures along all directions in parameter space. The second problem concerns the choice of direction. In particular, neither the component-wise choice (one parameter at a time) nor the random choice is suitable in strongly correlated cases. Using such choices results in highly autocorrelated samples.

Our approach would be to target each of those two issues individually. The resulting algorithm, Ensemble Slice Sampling (ESS), is invariant under affine transformations of the parameter space, meaning that its performance is not sensitive to linear correlations. Furthermore, ESS minimizes the computational cost of finding the slice by adaptively tuning the initial length scale. Last but not least, unlike most MCMC methods, ESS is trivially parallelizable, thus enabling the data analyst to take advantage of modern high performance computing facilities with multiple CPUs.

\subsection{Adaptively tuning the length scale}
\label{sec:approx}

Let us first consider the effect of the initial length scale on the performance of the univariate slice sampling method. For instance, if the initial length scale is $\lambda$ times smaller than the actual size of the slice, then the stepping-out procedure would require $\mathcal{O}(\lambda)$ steps in order to fix this. However, in this case, since the final interval is an accurate approximation of the slice there would probably be no contractions during the shrinking phase. On the other hand, when the initial length scale is larger than the actual slice then the number of expansions would be either one or zero. In this case though, there would be a number of contractions.\\

\noindent\textbf{Stochastic approximation:} As the task is to minimize the total number of expansions and contractions we employ and adapt the \emph{Robbins--Monro} stochastic approximation algorithm \citep{robbins1951stochastic} of \citet{tibbits2014automated}. Ideally, based on the reasoning of the previous paragraph, only one expansion and one contraction will take place. Therefore, the target ratio of number of expansions to total number of expansions and contractions is $1/2$. To achieve this, we update the length scale $\mu$ based on the following recursive formula:
\begin{equation}
    \label{eq:approx}
    \mu^{(t+1)} = 2 \mu^{(t)}\frac{N_{e}^{(t)}}{N_{e}^{(t)}+N_{c}^{(t)}}\, ,
\end{equation}
where $N_{e}^{(t)}$ and $N_{c}^{(t)}$ are the number of expansions and contractions during iteration $t$. It is easy to see that when the fraction $N_{e}^{(t)}/(N_{e}^{(t)}+N_{c}^{(t)})$ is larger than $1/2$ the length scale $\mu$ will be increased. In the case where the fraction is smaller than $1/2$ the length scale $\mu$ will be decreased accordingly. The optimization can stop either when the fraction is close to $1/2$ within a threshold or when a maximum number of tuning steps has been completed. The pseudocode for the first case is shown in \Algo{approximate}. In order to preserve detailed balance it is important to be sure that the adaptation stops after a finite number of iterations. In practice this happens after $\mathcal{O}(10)$ iterations. An alternative would be to use diminishing adaptation \citep{roberts2007coupling} but we found that our method is sufficient in practice (see Section 4.3 for more details).

\begin{algorithm}
\caption{Function to tune the length scale $\mu$.}
    \algolabel{approximate}
\begin{algorithmic}[1]
\STATE{\textbf{function} TuneLengthScale($t$, $\mu^{(t)}$, $N_{e}^{(t)}$,  $N_{c}^{(t)}$, $M^{\text{adapt}}$)}
\IF{$t\leq M^{\text{adapt}}$}
    \STATE{Compute $\mu^{(t+1)}$ using Equation \ref{eq:approx},}
    \STATE{\bf{return} $\mu^{(t+1)}$}
\ELSE
    \STATE{\bf{return} $\mu^{(t)}$}
\ENDIF
\end{algorithmic}
\end{algorithm}

\subsection{The choice of direction \& parallelization}
\label{sec:direction}

In cases where the parameters are correlated we can accelerate mixing by moving more frequently along certain directions in parameter space. One way of achieving this is to exploit some prior knowledge about the covariance of the target distribution. However, such an approach would either require significant hand-tuning or noisy estimations of the sample covariance matrix during an initial run of the sampler. For that reason we employ a different approach to exploit the covariance structure of the target distribution and preserve the hand-tuning-free nature of the algorithm.\\

\noindent\textbf{Ensemble of walkers:} Following the example of \cite{goodman2010ensemble} we define an ensemble of parallel chains, called walkers. In our case though, each walker is a individual slice sampler. The sampling proceeds by moving one walker at a time by slice sampling along a direction defined by a subset of the rest of walkers of the ensemble. As long as the aforementioned direction does not depend on the position of the current walker, the resulting algorithm preserves the detailed balance of the chain. Moreover, assuming that the distribution of the walkers resembles the correlated target distribution, the chosen direction will prefer directions of correlated parameters. 

We define an ensemble of $N$ parallel walkers as the collection $S = \lbrace \mathbf{X_{1}}, \dots,  \mathbf{X_{N}}\rbrace$. The position of each individual walker $\mathbf{X_{k}}$ is a vector $\mathbf{X_{k}}\in \mathbb{R}^{D}$ and therefore we can think of the ensemble $S$ as being in $\mathbb{R}^{N D}$. Assuming that each walker is drawn independently from the target distribution with density $p$, then the target distribution for the ensemble would be the product
\begin{equation}
    \label{eq:product}
    P(\mathbf{X_{1}}, \dots,  \mathbf{X_{N}}) = \prod_{k=1}^{N}p(\mathbf{X_{k}}) \,.
\end{equation}
The Markov chain of the ensemble would preserve the product density of equation \ref{eq:product} without the individual walker trajectories being Markov. Indeed, the position of $\mathbf{X_{k}}$ at iteration $t+1$ can depend on $\mathbf{X_{j}}$ at iteration $t$ with $j\neq k$.

Given the walker $\mathbf{X_{k}}$ that is to be updated there are arbitrary many ways off defining a direction vector from the complementary ensemble $S_{[k]}=\lbrace \mathbf{X_{j}},\: \forall j\neq k\rbrace$. Here we will discuss a few of them. Following the convention in the ensemble MCMC literature we call those recipes of defining direction vectors, \emph{moves}. Although the use of the ensemble might seem equivalent to that of a sample covariance matrix in the Adaptive Metropolis algorithm \citep{haario2001adaptive} the first has a higher information content as it encodes both linear and non-linear correlations. Indeed, having an ensemble of walkers allows for arbitrary many policies for choosing the appropriate directions along which the walkers move in parameter space. As we will shortly see, one of the choices (i.e. the Gaussian move, introduced later in this Section) is indeed the slice sampling analogue of a covariance matrix. However, other choices (i.e. Differential move or Global move) can take advantage of the non-Gaussian nature of the ensemble distribution and thus propose more informative moves. As it will be discussed later in this section, those advanced moves make no assumption of Gaussianity for the target distribution. Furthermore, as we will show in the last part of this section, the ensemble can also be easily parallelised. \\

\begin{algorithm}
\caption{Function to return a differential move direction vector.}
    \algolabel{differential}
\begin{algorithmic}[1]
\STATE{\textbf{function} DifferentialMove($k$, $\mu$, $S$)}
\STATE{Draw two walkers $\mathbf{X_{l}}$, and $\mathbf{X_{m}}$ uniformly and without replacement from the complementary ensemble $S$},
\STATE{Compute direction vector $\bm{\eta}_{k}$ using Equation \ref{eq:diff},}
\STATE{\bf{return} $\bm{\eta}_{k}$}
\end{algorithmic}
\end{algorithm}

\noindent\textbf{Affine transformations and invariance}: Affine invariance is a property of certain MCMC samplers first introduced in the MCMC literature by \citet{goodman2010ensemble}. An MCMC algorithm is said to be affine invariant if its performance is invariant under the bijective mapping $g:\mathbb{R}^{D}\rightarrow \mathbb{R}^{D}$ of the form $\mathbf{Y}=A \mathbf{X} + b$ where $A\in \mathbb{R}^{D\times D}$ is a matrix and $b\in\mathbb{R}^{D}$ is a vector. Linear transformations of this form are called affine transformations and describe rotations, rescaling along specific axes as well as translations in parameter space. Assuming that $\mathbf{X}$ has the probability density $p(\mathbf{X})$, then $\mathbf{Y}=A\mathbf{X}+b$ has the probability density
\begin{equation}
    \label{eq:affinedensity}
    p_{A,b}(\mathbf{Y})=p(A\mathbf{X}+b)\propto p(\mathbf{X})\,.
\end{equation}

Given a density $p$ as well as an MCMC transition operator $\mathcal{T}$ such that $\mathbf{X}(t+1) = \mathcal{T} \big(\mathbf{X}(t);p\big)$ for any iteration $t$ we call the operator $\mathcal{T}$ affine invariant if
\begin{equation}
    \label{eq:invariance}
    \mathcal{T}\big(A\mathbf{X}+b;p_{A,b}\big) = A\:\mathcal{T}\big(\mathbf{X};p\big)+b
\end{equation}
for $\forall A\in \mathbb{R}^{D\times D}$ and $\forall b \in \mathbb{R}^{D}$. In case of an ensemble of walkers we define an affine transformation from $\mathbb{R}^{N D}$ to $\mathbb{R}^{N D}$ as
\begin{equation}
    \label{eq:transformensemble}
    S = \lbrace \mathbf{X_{1}}, \dots, \mathbf{X_{N}}\rbrace \xrightarrow{A, b} \lbrace A \mathbf{X_{1}}+b, \dots, A\mathbf{X_{N}}+b\rbrace \,.
\end{equation}

The property of affine invariance is of paramount importance for the development of efficient MCMC methods. As we have discussed already, proposing samples more frequently along certain directions can accelerate sampling by moving further away in parameter space. Given that most realistic applications are highly skewed or anisotropic and are characterised by some degree of correlation between their parameters, affine invariant methods are an obvious choice of a tool that can be used in order to achieve high levels of efficiency.
\\

\begin{figure}[t!]
    \centering
    \includegraphics[scale=0.55]{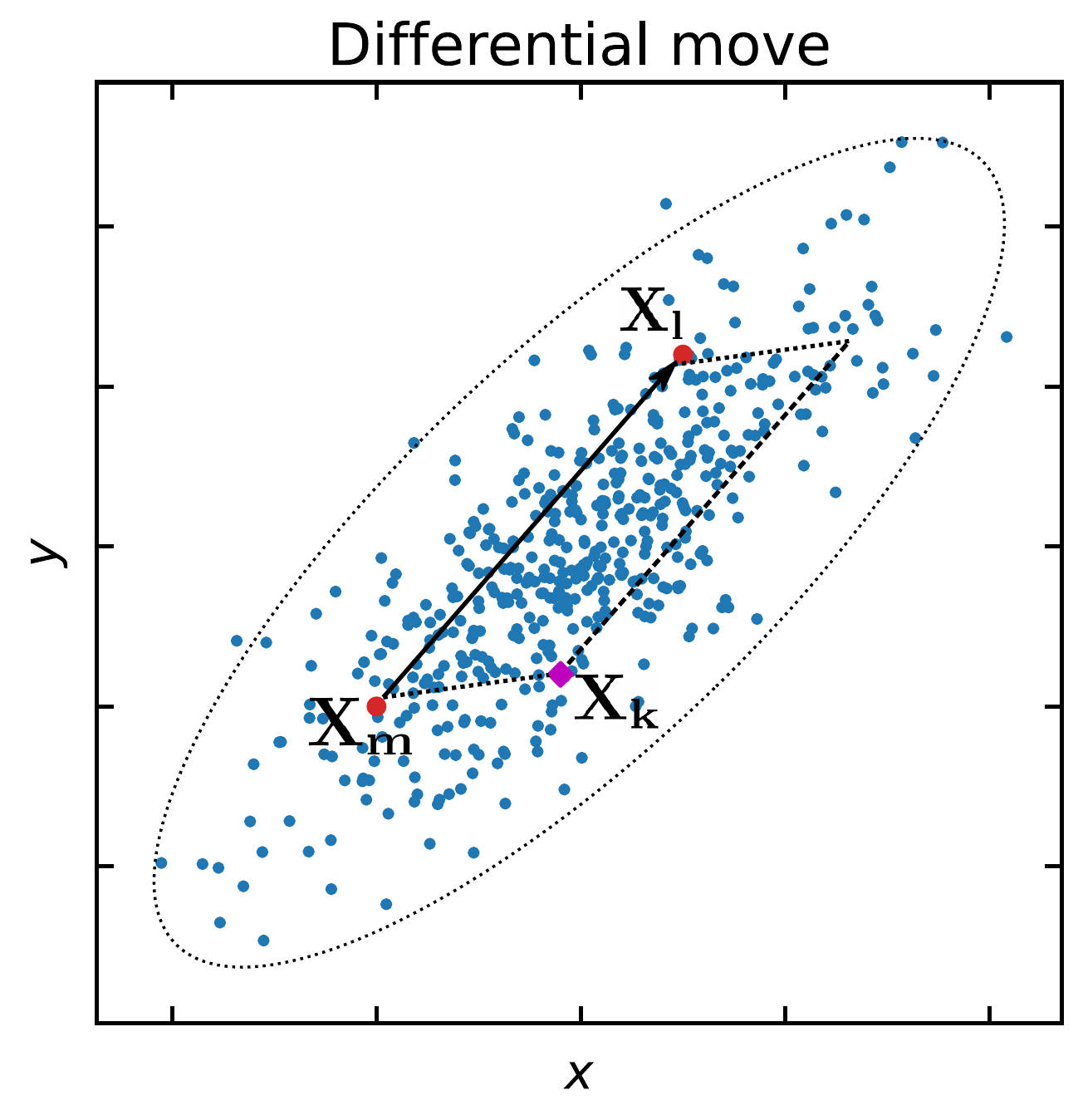}
    \caption{The plot shows the differential direction move. Two walkers (red) are uniformly sampled from the complementary ensemble (blue). Their positions define the direction vector (solid black). The selected walker (magenta) then moves by Slice Sampling along the parallel direction (dashed black).}
\label{fig:diff}
\end{figure}

\noindent\textbf{Differential move:} The differential direction choice works by moving the walker $\mathbf{X}_{k}$ based on two randomly chosen walkers $\mathbf{X}_{l}$ and $\mathbf{X}_{m}$ of the complementary ensemble $S_{[k]}=\lbrace \mathbf{X_{j}},\: \forall j\neq k\rbrace$ \citep{gilks1994adaptive}, see Figure \ref{fig:diff} for a graphical explanation. In particular, we move the walker $\mathbf{X}_{k}$ by slice sampling along the vector $\bm{\eta}_{k}$ defined by the difference between the walkers $\mathbf{X}_{l}$ and $\mathbf{X}_{m}$. It is important to notice here that the vector $\bm{\eta}_{k}$ is not a unit vector and thus carries information about both the length scale and the optimal direction of movement. It will also prove to be more intuitive to include the initial length scale $\mu$ in the definition of the direction vector in the following way:
\begin{equation}
    \label{eq:diff}
    \bm{\eta}_{k}= \mu \big( \mathbf{X}_{l}-\mathbf{X}_{m}\big)\, .
\end{equation}

The pseudocode for a function that, given the value of $\mu$ and the complementary ensemble $S$, returns a differential direction vector $\bm{\eta}_{k}$ is shown in \Algo{differential}. Furthermore, the Differential move is clearly affine invariant. Assuming that the distribution of the ensemble of walkers follows the target distribution and the latter is highly elongated or stretched along a certain direction then the proposed direction given by equation \ref{eq:diff} will share the same directional asymmetry.   \\

\noindent\textbf{Gaussian move:} The direction vector $\bm{\eta}_{k}$ can also be drawn from a normal distribution with the zero mean and the covariance matrix equal to the sample covariance of the complementary ensemble $S_{[k]}$,
\begin{equation}
    \label{eq:cov}
    \mathbf{C}_{S}=\frac{1}{|S|}\sum_{j\in S}\big(\mathbf{X}_{j}-\bar{\mathbf{X}}_{S}\big)\big(\mathbf{X}_{j}-\bar{\mathbf{X}}_{S}\big)^{t}\, .
\end{equation}
We chose to include the initial length scale $\mu$ in this definition as well:
\begin{equation}
    \label{eq:gaussian}
    \frac{\bm{\eta}_{k}}{2\mu} \sim \mathcal{N}\big(\mathbf{0},\mathbf{C}_{S} \big)\, .
\end{equation}
The factor of $2$ is used so that the magnitude of the direction vectors are consistent with those sampled using the differential direction choice in the case of Gaussian-distributed walkers.

\begin{figure}[t!]
    \centering
    \includegraphics[scale=0.55]{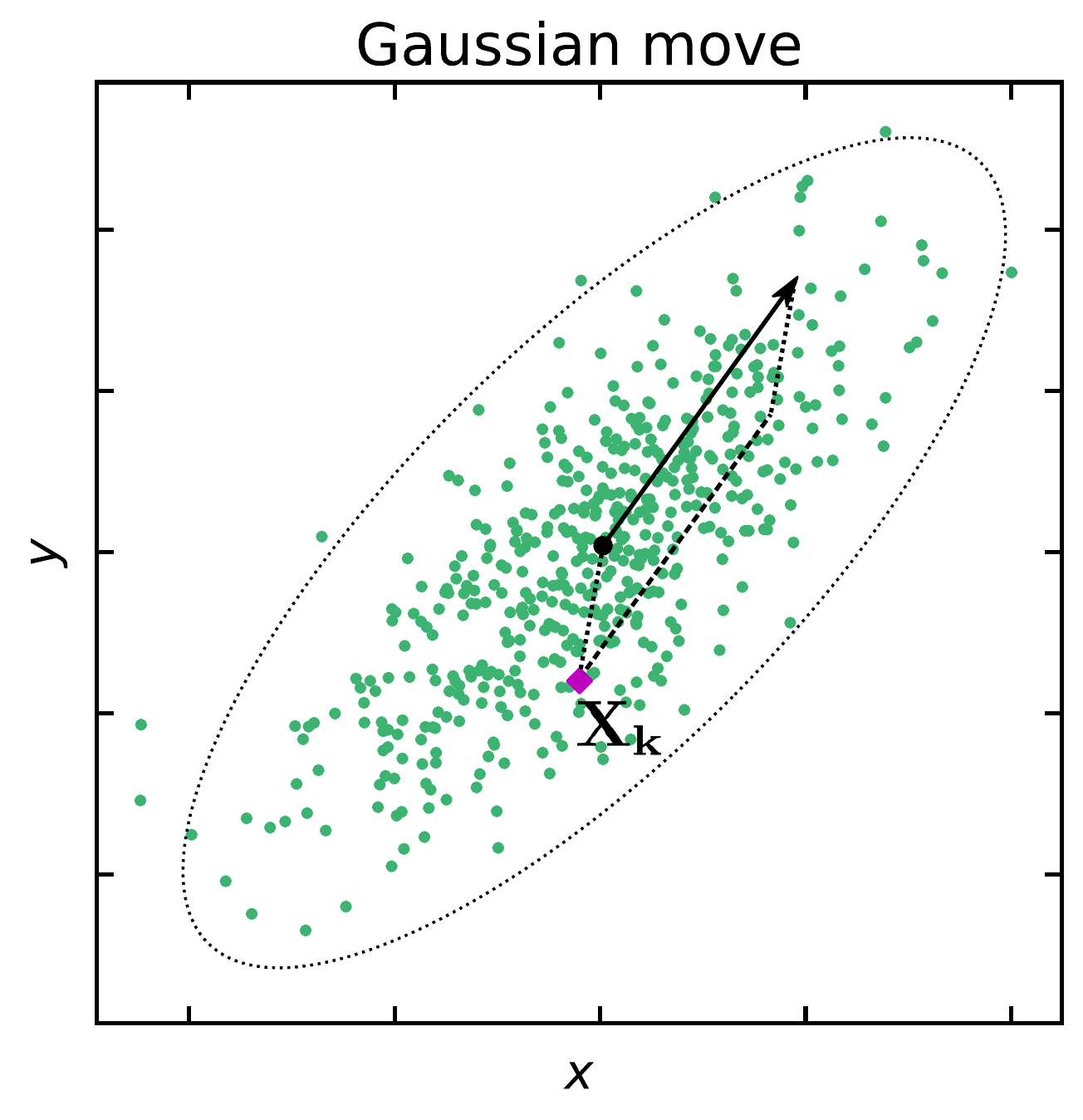}
    \caption{The plot shows the Gaussian direction move. A direction vector (solid black) is sampled from the Gaussian-approximated distribution of the walkers of the complementary ensemble (green). The selected walker (magenta) then moves by Slice Sampling along the parallel direction (dashed black).}
\label{fig:gauss}
\end{figure}

The pseudocode for a function that, given the value of $\mu$ and the complementary ensemble $S$, returns a Gaussian direction vector $\bm{\eta}_{k}$ is shown in \Algo{gaussian}. See Figure \ref{fig:gauss} for a graphical explanation of the method. Moreover, just like the Differential move, the Gaussian move is also affine invariant. In the limit in which the number of walkers is very large and the target distribution is normal, the first reduces to the second. Alternatively, assuming that the distribution of walkers follows the target distribution then the covariance matrix of the ensemble would be the same as that of independently drawn samples from the target density. Therefore any anisotropy characterising the target density would also be present in the distribution of proposed directions given by equation \ref{eq:gaussian}.
\begin{algorithm}
\caption{Function to return a Gaussian Move direction vector.}
    \algolabel{gaussian}
\begin{algorithmic}[1]
\STATE{\textbf{function} GaussianMove($k$, $\mu$, $S$)}
\STATE{Estimate sample covariance $\mathbf{C}_{S}$ of the walkers in the complementary ensemble $S$ using Equation \ref{eq:cov},}
\STATE{Sample $\bm{\eta}_{k}/(2\mu)\sim \mathcal{N}\big(\mathbf{0},\mathbf{C}_{S} \big)$,}
\STATE{\bf{return} $\bm{\eta}_{k}$}
\end{algorithmic}
\end{algorithm}

\noindent\textbf{Global move:} ESS and its variations described so far (i.e. differential move, Gaussian move) have as much difficulty traversing the low probability regions between modes/peaks in multimodal distributions as most local MCMC methods (e.g. Metropolis, Hamiltonian Monte Carlo, Slice Sampling, etc.). Indeed, multimodal distributions are often the most challenging cases to sample from. Fortunately, Ensemble Slice Sampling's flexibility allows to construct advanced moves which are specifically designed to handle multimodal cases even in moderate to high dimensional parameter spaces. The \emph{global move} is such an example.

We first fit a \emph{Gaussian Mixture} to the distribution of the walkers of the complementary ensemble $S_{[k]}$ using \emph{Variational Inference}. To avoid defining the number of components of the Gaussian Mixture we use a \emph{Dirichlet process} as the prior distribution for the Gaussian Mixture weights\footnote{To this end we use the Scikit-Learn implementation of the Dirichlet process Gaussian mixture.} \citep{gorur2010dirichlet}. The exact details of the construction of the Dirchlet process Gaussian mixture (DPGM) are beyond the scope of this work and we direct the reader to \citet{gorur2010dirichlet} and \citet{bishop2006pattern} for more details. One of the major benefits of fitting the DPGM using variational inference compared to the expectation--maximisation (EM) algorithm~\citep{dempster1977maximum} that is often used is the improved stability. In particular, the use of priors in the variational Bayesian treatment guarantees that Gaussian components do not collapse into specific data points. This regularisation due to the priors leads to component covariance matrices that do not diverge even when the number of data points (i.e. walkers in our case) in a component is lower than the number of dimensions. In our case, this  means that even if the number of walkers located in a mode of the target distribution is small DPGM would still identify that mode correctly. In such cases, the covariance of the  component that corresponds to that mode would be over--estimated. This however does not affect the performance of the Global move as the latter does not rely on exact estimates of the component covariance matrices.\footnote{Indeed the covariance matrix of a component only enters through equation \ref{eq:globalB} but then it is re--scaled by the factor $\gamma$.}

In practice, we recommend using more than the minimum number of walkers in cases of multimodal distributions (e.g. at least two times as many in bimodal cases). We found that the computational overhead introduced by the variational fitting of the DPGM is negligible compared to the computational cost of the evaluation of the model and posterior distribution in common problems in physics, astrophysics and cosmology. Indeed the cost is comparable, and only a few times higher than the Differential or Gaussian move. The reason for that is the relatively small number of walkers (i.e. $\mathcal{O}(10-10^3)$) that simplifies the fitting procedure.

Once fitting is done, we have a list of the means and covariance matrices of the components of the Gaussian Mixture. As the ensemble of walkers traces the structure of the target distribution, we can use the knowledge of the means and covariance matrices of the Gaussian Mixture to construct efficient direction vectors. Ideally, we prefer direction vectors that connect different modes. This way, the walkers will be encouraged to move along those directions that would otherwise be very unlikely to be chosen.

We uniformly select two walkers of the complementary ensemble and identify the Gaussian components to which they belong, say $i$ and $j$. There are two distinct cases and we will treat them as such. In case A, $i = j$, meaning that the selected walkers originate from the same component. In case B, $i \neq j$, meaning that the two walkers belong to different components and thus probably different peaks of the target distribution. 

As we will show next, only in case B, we can define a direction vector that favors mode-jumping behaviour. In case A, we can sample a direction vector from the Gaussian component that the two select walkers belong to\footnote{In practice we use uniformly sample two walkers from the list of walkers that DPGM identified in that mode. This step removes any dependency on covariance matrix estimates.}:
\begin{equation}
    \label{eq:globalA}
    \frac{\bm{\eta}_{k}}{2\mu} \sim \mathcal{N}\big( \bm{0}, \bm{C}_{i=j} \big)\, ,
\end{equation}
where $\bm{C}_{i=j}$ is the covariance matrix of the i$_{\rm th}$ (or equivalently j$_{\rm th}$) component. Just as in the Gaussian move, the mean of the proposal distribution is zero so that we can interpret $\bm{\eta}$ as a direction vector.

In case B, where the two selected walkers belong to different components, $i \neq j$, we will follow a different procedure to facilitate long jumps in parameter space. We will sample two vectors, one from each component:
\begin{equation}
    \label{eq:globalB}
    \bm{\eta}_{k, n} \sim \mathcal{N}\big( \bm{\mu}_{n}, \gamma \bm{C}_{n} \big)\, ,
\end{equation}
for $n=i$ or $n=j$. Here, $\bm{\mu}_{n}$ is the mean of the nth component and $\bm{C}_{n}$ is its covariance matrix. In practise, we also re-scale the covariance by a factor of $\gamma = 0.001$, which results in direction vectors with lower variance in their orientation. $\gamma < 1$ ensures that the chosen direction vector is close to the vector connecting the two peaks of the distribution. Finally, the direction vector will be defined as:
\begin{equation}
    \label{eq:global}
    \bm{\eta}_{k} = 2 \big( \bm{\eta}_{k,i} - \bm{\eta}_{k,j} \big)\, .
\end{equation}
The factor of $2$ here is chosen to better facilitate mode-jumping. There is also no factor of $\mu$ in the aforementioned expression, since in this case there is no need for the scale factor to be tuned.

\begin{figure}[t!]
    \centering
    \includegraphics[scale=0.55]{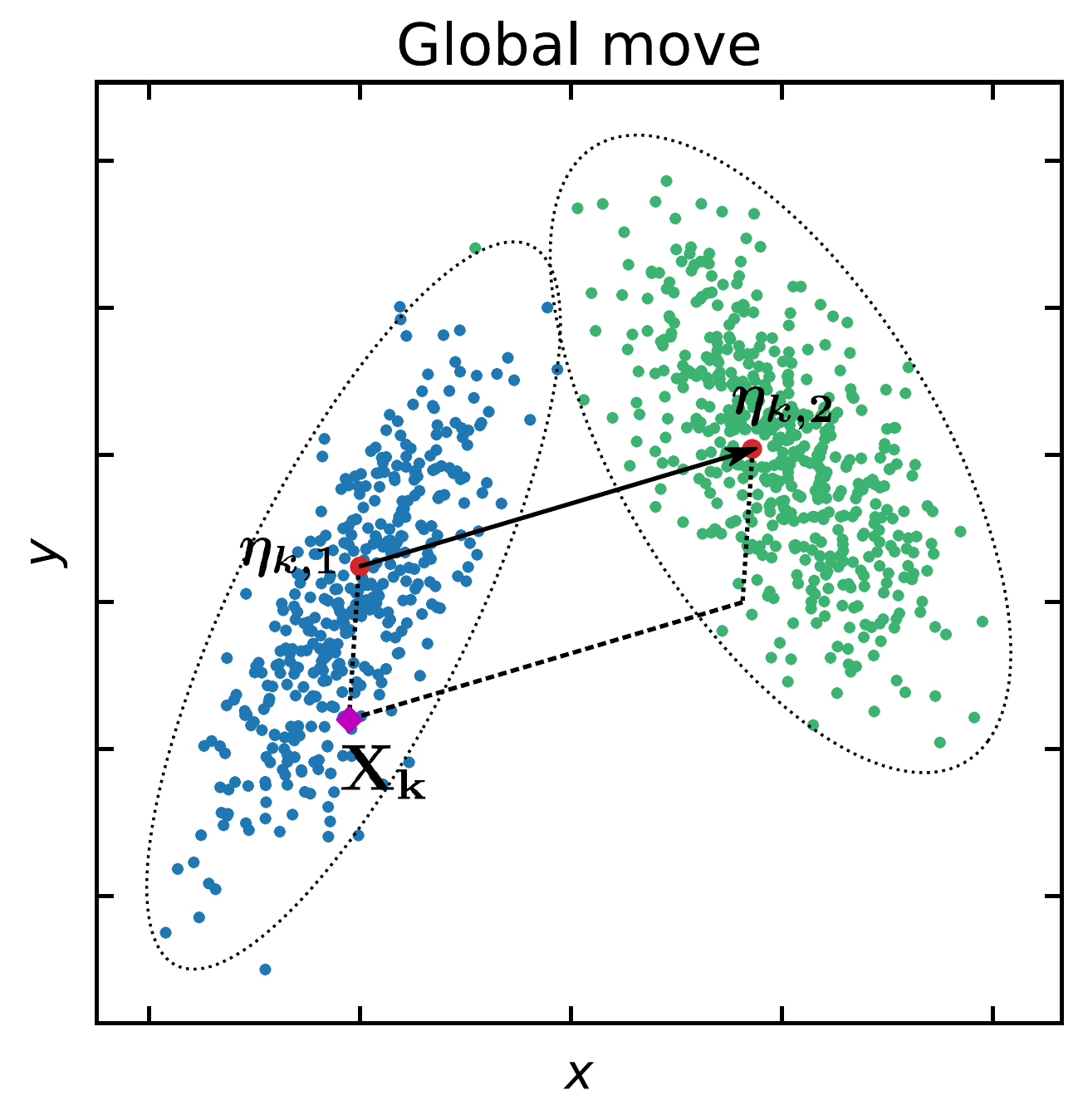}
    \caption{The plot shows the global direction move assuming that the uniformly selected pair of walkers of the complementary ensemble belongs to different components (blue and green). A position (red) is sampled from each component (using the re-scaled by $\gamma$ covariance matrix). Those two points (red) define the direction vector (black) connecting the two modes (blue and green). The selected walker (magenta) then moves by slice sampling along the parallel direction (dashed).}
\label{fig:global}
\end{figure}

The pseudocode for a function that, given the complementary ensemble $S$, returns a Global direction vector $\bm{\eta}_{k}$ is shown in \Algo{global}. See Figure \ref{fig:global} for a graphical explanation of the method. It should be noted that for the global move to work at least one walker needs to be present on each well separated mode.
\begin{algorithm}
\caption{Function to return a global move direction vector.}
    \algolabel{global}
\begin{algorithmic}[1]
\STATE{\textbf{function} GlobalMove($k$, $\mu$, $S$)}
\STATE{Fit Dirichlet process Gaussian mixture (DPGM) to the complementary ensemble $S_{[k]}$,}
\STATE{If $N$ is the number of components of the DPGM then select two components $i, j$ uniformly such that $i \neq j$,}
\IF{$i = j$}
    \STATE{Sample $\bm{\eta}_{k}/(2\mu) \sim \mathcal{N}\big( \bm{0}, \bm{C}_{i=j} \big)$,}
\ELSE
    \STATE{Sample $\bm{\eta}_{k, n} \sim \mathcal{N}\big( \bm{\mu}_{n}, \gamma \bm{C}_{n} \big)$ for $n=i, j$,}
    \STATE{Compute direction vector $\bm{\eta}_{k}$ using Equation \ref{eq:global},}
\ENDIF
\STATE{\bf{return} $\bm{\eta}_{k}$}
\end{algorithmic}
\end{algorithm}

Here we introduced three general and distinct moves that can be used in a broad range of cases. In general, the global move requires a higher number of walkers than the differential or Gaussian move in order to perform well. We found that the differential and Gaussian moves are good choices for most target distributions whereas the global move is only necessary in highly dimensional and multimodal cases. One can use the information in the complementary ensemble to construct more moves tailor-made for specific problems. Such additional moves might include Kernel Density Estimation or Clustering methods and as long as the information used comes from the complementary ensemble (and not from the walker that would be updated) the detailed balance is preserved.\\

\noindent\textbf{Parallelizing the ensemble:} Instead of evolving the ensemble by moving each walker in turn we can do this in parallel. A naive implementation of this would result in a subtle violation of detailed balance. We can avoid this by splitting the ensemble into two sets of walkers \citep{foreman2013emcee} of $n_{\text{Walkers}} / 2$ each. We can now update the positions of all the walkers in the one set in parallel along directions defined by the walkers of the other set (the complementary ensemble). Then we can perform the same procedure for the other set. In accordance with equation \ref{eq:product}, the stationary distribution of the split ensemble would be
\begin{equation}
    \label{eq:split}
    P(\mathbf{X_{1}},\dots,\mathbf{X_{N}}) = \prod_{k=1}^{N/2}p(\mathbf{X_{k}}) \prod_{k=1+N/2}^{N}p(\mathbf{X_{k}})\,.
\end{equation}
The method generates samples from the target distribution by simulating a Markov chain which leaves this product distribution invariant. The transition operator $\mathcal{T}_{1}$ that updates the walkers of the first set (i.e. $k=1,\dots,N/2$) uses the walkers of the complementary ensemble (i.e. $k=1+N/2,\dots,N$) and vice versa for the transition operator $\mathcal{T}_{2}$ that acts on the second set. In the context of ESS the aforementioned transition operators correspond to a single iteration of \Algo{final} coupled with one of the moves (e.g. Differential move).

It follows from the ensemble splitting technique that the maximum number of CPUs used without any of them being idle is equal to the total number of walkers updated concurrently, that is $n_{\text{Walkers}} / 2$.  We will also verify this empirically in Section \ref{sec:empirical}. Of course, this does not mean that if there are more CPUs available they cannot be used as we can always increase the size of the ensemble to match the available CPUs.

Combining this technique with the stochastic approximation solution of Subsection \ref{sec:approx} and the choices (moves) of direction and ensemble-splitting technique of this subsection leads to the Ensemble Slice Sampling method of \Algo{final}\footnote{Perhaps a small detail, but we have included the length scale in the definition of the direction vector $\mathbf{\eta}$ and therefore it does not appear in the definition of the $(L, R)$ interval.}. Of course, another move (e.g. Gaussian, global) can be used instead of the differential move in \Algo{final}. Finally, the minimum number of walkers used should be twice the number of parameters. Using fewer walkers than that could lead to erroneous sampling from a lower dimensional parameter space \citep{ter2006markov}.

In general, parallelizing a slice sampler is not trivial (e.g. as it is for Metropolis) because each update requires an unknown number of probability density evaluations. However, because of the affine invariance (i.e. performance unaffected by linear correlations) induced by the existence of the ensemble, all iterations require on average the same number of probability density evaluations (i.e. usually $5$ if the stochastic approximation for the length scale $\mu$ is used). Therefore, the parallelization of Ensemble Slice Sampling is very effective in practice. Furthermore, the benefit of having parallel walkers instead of parallel independent chains (e.g. such as in Metropolis sampling) is clear, the walkers share information about the covariance structure of the distribution thus accelerating mixing.

\begin{algorithm}[ht!]
\caption{Single Iteration $t$ of Ensemble Slice Sampling.}
    \algolabel{final}
\begin{algorithmic}[1]
\STATE{Given $t$, $f$, $\mu^{(t)}$, $S_{[0]}$, $S_{[1]}$, $M^{\rm adapt}$:}
\STATE{Initialise $N_{e}^{(t)} = 0$ and $N_{c}^{(t)} = 0$},
\FOR{$i=0, 1$}
    \FOR{$k=1, ..., N/2$}
        \STATE{$k \leftarrow k + i N/2$}
        \STATE{Compute direction vector $\bm{\eta}_{k}\leftarrow$ DifferentialMove($k$, $\mu^{(t)}$, $S_{[i]}$)}
        \STATE{Sample $Y \sim \text{Uniform}(0,f(\mathbf{X_{k}}^{(t)}))$}
        \STATE{Sample $U \sim \text{Uniform}(0,1)$}
        \STATE{Set $L \leftarrow - U$, and $R \leftarrow L + 1$}
        \WHILE{$Y < f(L)$}
            \STATE{$L \leftarrow L - 1$}
            \STATE{$N_{e}^{(t)} \leftarrow N_{e}^{(t)} + 1$}
        \ENDWHILE
        \WHILE{$Y < f(R)$}
            \STATE{$R \leftarrow R + 1$}
            \STATE{$N_{e}^{(t)} \leftarrow N_{e}^{(t)} + 1$}
        \ENDWHILE
        \WHILE{True}
            \STATE{Sample $X' \sim \text{Uniform}(L,R)$}
            \STATE{Set $Y' \leftarrow f(X'\bm{\eta}_{k} + \mathbf{X_{k}}^{(t)})$}
            \IF{$Y<Y'$}
                \STATE{\bf{break}}
            \ENDIF
            \IF{$X'<0$}
                \STATE{$L \leftarrow X'$}
                \STATE{$N_{c}^{(t)} \leftarrow N_{c}^{(t)} + 1$}
            \ELSE
                \STATE{$R \leftarrow X'$}
                \STATE{$N_{c}^{(t)} \leftarrow N_{c}^{(t)} + 1$}
            \ENDIF
        \ENDWHILE
        \STATE{Set $\mathbf{X_{k}}^{(t+1)} \leftarrow X' \bm{\eta}_{k} + \mathbf{X_{k}}^{(t)}$}
    \ENDFOR
\ENDFOR

\STATE{$\mu^{(t+1)} \leftarrow$ TuneLengthScale($t$, $\mu^{(t)}$, $N_{e}^{(t)}$,  $N_{c}^{(t)}$, $M^{adapt}$),}
\end{algorithmic}
\end{algorithm}

\section{Empirical evaluation}
\label{sec:empirical}

To empirically evaluate the sampling performance of the Ensemble Slice Sampling algorithm we perform a series of tests. In particular, we compare its ability to sample from two demanding target distributions, namely the \emph{autoregressive process of order 1} and the \emph{correlated funnel}, against the Metropolis and Standard Slice Sampling algorithms. The Metropolis' proposal scale was tuned to achieve the optimal acceptance rate, whereas the initial length scale of Standard Slice Sampling was tuned using the stochastic scheme of \Algo{approximate}. Ensemble Slice Sampling significantly outperforms both of them. These tests help establish the characteristics and advantages of Ensemble Slice Sampling. Since our objective was to develop a gradient-free black-box method we then proceed to compare Ensemble Slice Sampling with a list of gradient-free ensemble methods such as \emph{Affine Invariant Ensemble Sampling} (AIES), \emph{Differential Evolution Markov Chain} (DEMC) and \emph{Kernel Density Estimate Metropolis} (KM) on a variety of challenging target distributions. Moreover, we are also interested in assessing the convergence rate of the length scale $\mu$ during the first iterations as well as the parallel scaling of the method in the presence of multiple CPUs. Unless otherwise specified we use the differential move for the tests. Unlike ESS that has an acceptance rate of $1$, AIES's and DEMC's acceptance rate is related to the number of walkers. For that reason, and for the sake of a fair comparison, we made sure the selected number of walkers in all examples would yield the optimal acceptance rate for AIES and DEMC. As we will discuss further in Section \ref{sec:discussion} it makes sense to increase the number of walkers in cases of multimodal distributions or strong non-linear correlations. In general though, we recommend using the minimum number of walkers (i.e. twice the number of dimensions) as the default choice and increase it only if it is required by a specific application. For more rules and heuristics about the initialisation and number of walkers we direct the interested reader to Section \ref{sec:discussion}.

\subsection{Performance tests}

\noindent\textbf{Autoregressive process of order 1:} In order to investigate the performance of ESS. in high dimensional and correlated scenarios we chose a highly correlated Gaussian as the target distribution. More specifically, the target density is a discrete-time \emph{autoregressive process of order 1}, also known as AR(1). This particular target density is ideally suited for benchmarking MCMC algorithms since the posterior density in many scientific studies often approximates a correlated Gaussian. Apart from that, the AR(1) is commonly used as a prior for time-series analysis.

The AR(1) distribution of a random vector $\bm{X}=(X_{1},...,X_{N})$ is defined recursively as follows:
\begin{equation}
    \label{eq:ar1}
    \begin{split}
        X_{1} \sim &\;\mathcal{N}(0,1)\, , \\ 
        X_{2}|X_{1} \sim &\;\mathcal{N}(\alpha X_{1},\beta^{2})\, , \\
        &\vdots \\
        X_{N}|X_{N-1} \sim &\;\mathcal{N}(\alpha X_{N-1},\beta^{2})\, ,
    \end{split}
\end{equation}
where the parameter $\alpha$ controls the degree of correlation between parameters and we chose it to be $\alpha = 0.95$. We set $\beta = \sqrt{1-\alpha^{2}}$ so that the marginal distribution of all parameters is $\mathcal{N}(0,1)$. We also set the number of dimensions to $N=50$. 

\begin{figure*}[t!]
    \centering
    \includegraphics[width=.3\textwidth]{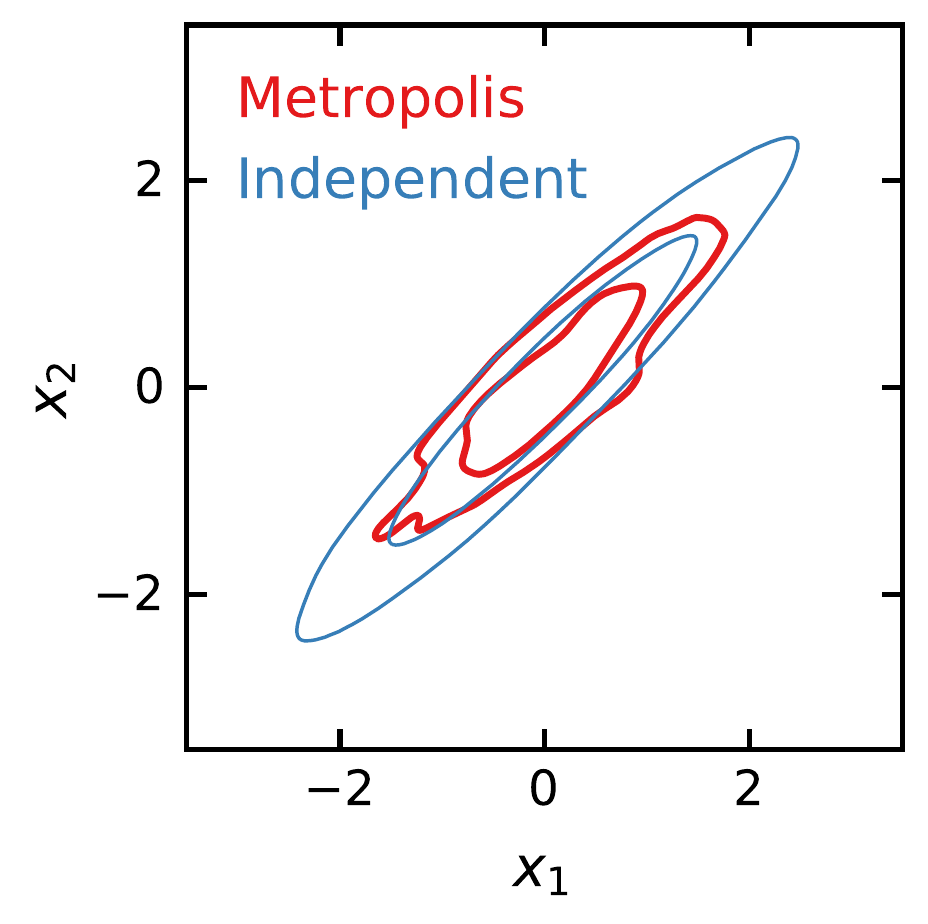}
    \includegraphics[width=.3\textwidth]{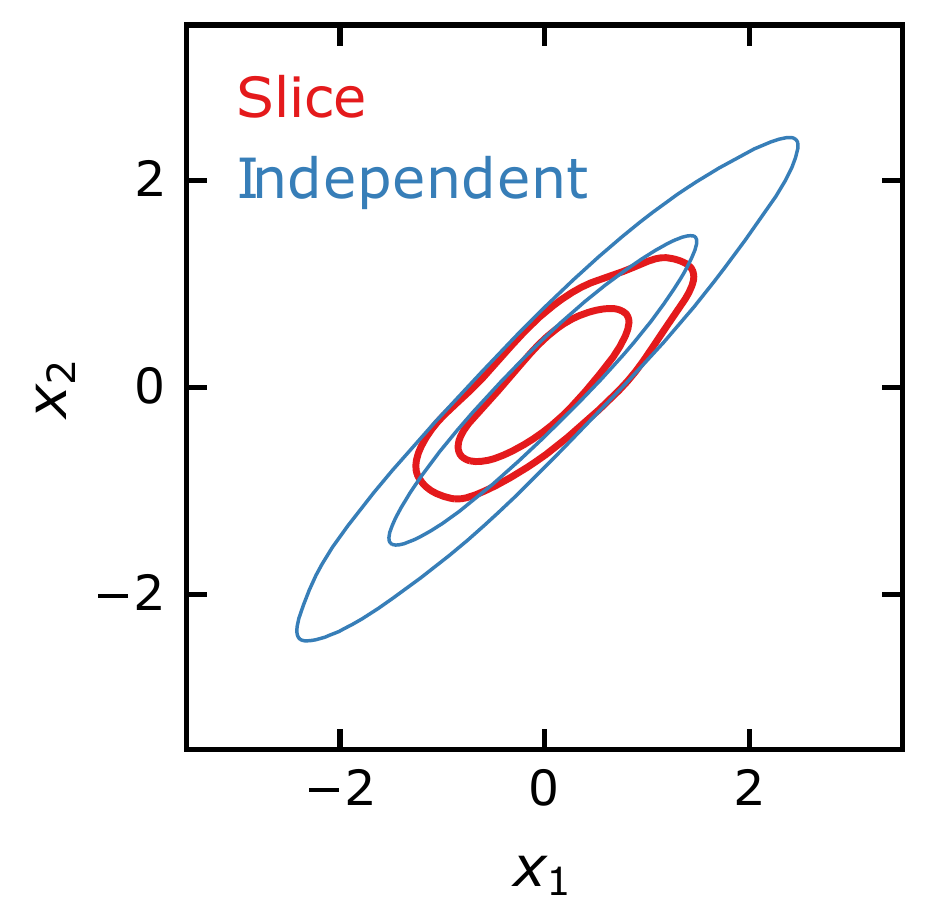}
    \includegraphics[width=.3\textwidth]{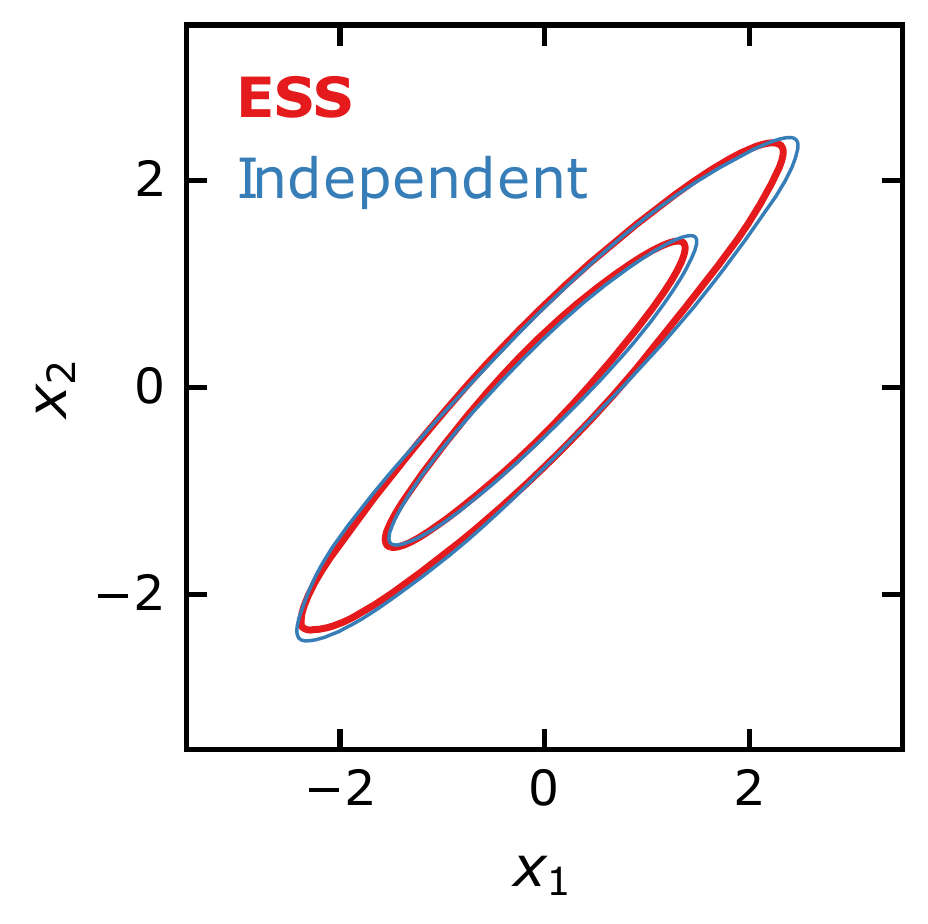}
\caption{The plots compare the 1-sigma and 2-sigma contours generated by the optimised random-walk Metropolis (left), Standard Slice (centre) and Ensemble Slice Sampling (right) methods to those obtained by Independent Sampling (blue) for the AR(1) distribution. All samplers used the same number of probability density evaluations, $3\times 10^{5}$. Only the first two dimensions are shown here.}
\label{fig:ar1}
\end{figure*}

\begin{table}[ht!]
    \centering
    \caption{The table shows a comparison of the optimally tuned Metropolis, Standard Slice, and Ensemble Slice Sampling with the differential move (ESS-D) and the Gaussian move (ESS-G) respectively in terms of the integrated autocorrelation time (IAT) and the number of effective samples per evaluation of the probability density (efficiency) multiplied by $10^4$. These metrics are formally defined in Appendix \ref{app:ess}. The target distributions are the 50--dimensional autoregressive process of order 1 and the 25--dimensional correlated funnel distribution. The total number of iterations was set to $10^{7}$.}
    \def\arraystretch{1.1}
    \begin{tabular}{lccccc}
        \toprule[0.75pt]
         & Metropolis   & Slice   & \textbf{ESS-D} & \textbf{ESS-G} \\
        \midrule[0.5pt]
        \multicolumn{4}{l}{Autoregressive process of order 1} \\
        \midrule[0.5pt]
        IAT          &    4341    &    2075    &   $\mathbf{111}$ &   $\mathbf{107}$   \\
        efficiency   &    2.3    &    1.0    & $\mathbf{17.5}$ & $\mathbf{17.8}$  \\
        \midrule[0.5pt]
        \multicolumn{4}{l}{Correlated funnel distribution} \\
        \midrule[0.5pt]
        IAT          &    -    &    3905    &   $\mathbf{129}$ &   $\mathbf{141}$   \\
        efficiency   &    -    &    0.5    &  $\mathbf{15.3}$ &  $\mathbf{14.0}$ \\
        \bottomrule[0.75pt]
        \end{tabular}
    \label{tab:table1}
\end{table}

For each method, we measured the mean \emph{integrated autocorrelation time} (IAT), and the number of effective samples per evaluation of the probability density function, also termed \emph{efficiency} (see Appendix \ref{app:ess} for details). For this test we ran the samplers for $10^{7}$ iterations. In this example we used the minimum number of walkers (i.e. 100 walkers) for ESS and the equivalent number of probability evaluations for Metropolis and Slice Sampling with each walker initialised at a position sampled from the distribution $\mathcal{N}(0,1)$. The results are presented in Table \ref{tab:table1}. The chain produced by Ensemble Slice Sampling has a significantly shorter IAT ($20-40$ times) compared to either of the other two methods. Furthermore, Ensemble Slice Sampling, with either Differential or Gaussian move, generates an order of magnitude greater number of independent samples per evaluation of the probability density. In this example the Differential and Gaussian moves have achieved almost identical IAT values and efficiencies.

To assess the mixing rate of Ensemble Slice Sampling, we set the maximum number of probability density evaluations to $3\times 10^{5}$ and show the results in Figure \ref{fig:ar1}. We compare the results of Ensemble Slice Sampling with those obtained via the optimally tuned Metropolis and Standard Slice Sampling methods. Ensemble Slice Sampling significantly outperforms both of them, being the only one with a chain resembling the target distribution in the chosen number of probability evaluations.\\

\noindent\textbf{Correlated funnel:} The second test involves a more challenging distribution, namely the correlated funnel distribution adapted from \citet{neal2003slice}. The funnel, tornado like, structure is common in Bayesian hierarchical models and possesses characteristics that render it a particularly difficult case. The main difficulty originates from the fact that there is a region of the parameter space where the volume of the region is low but the probability density is high, and another region where the opposite holds. 

Suppose we want to sample an N--dimensional vector $\bm{X}=(X_{1},...,X_{N})$ from the correlated funnel distribution. The marginal distribution of $X_{1}$ is Gaussian with mean zero and unit variance. Conditional on a value of $X_{1}$, the vector $\bm{X}_{2-N}=(X_{2},...,X_{N})$ is drawn from a Gaussian with mean zero and a covariance matrix in which the diagonal elements are $\exp(X_{1})$, and the non-diagonal equal to $\gamma\exp(X_{1})$. If $\gamma=0$, the parameters $X_{2}$ to $X_{N}$ conditional on $X_{1}$ are independent and the funnel distribution resembles the one proposed by \citet{neal2003slice}. The value of $\gamma$ controls the degree of correlation between those parameters. When $\gamma = 0$ the parameters are uncorrelated. For the following test we chose this to be $\gamma = 0.95$.  We set the number of parameters $N$ to $25$.

Using $10^{7}$ iterations, we estimated the IAT and the efficiency of the algorithms for this distribution as shown in Table \ref{tab:table1}. Just like in the AR(1) case we used the minimum number (i.e. 50) of walkers for ESS with each walker initialised at a position sampled from the distribution $\mathcal{N}(0,1)$. Since the optimally-tuned Metropolis fails to sample from this particular distribution, we do not quote any results. The Metropolis sampler is unable to successfully explore the region of parameter space with negative $X_{1}$ values. The presence of strong correlations renders the Ensemble Slice Sampler $30$ times more efficient than the Standard Slice Sampling algorithm on this particular example. In this example, the Differential move outperforms the Gaussian move in terms of efficiency, albeit by a small margin. In general, we expect the former to be more flexible than the latter since it makes no assumption about the Gaussianity of the target-distribution and recommend it as the default configuration of the algorithm.

\begin{figure*}[t!]
    \centering
    \includegraphics[width=.3\textwidth]{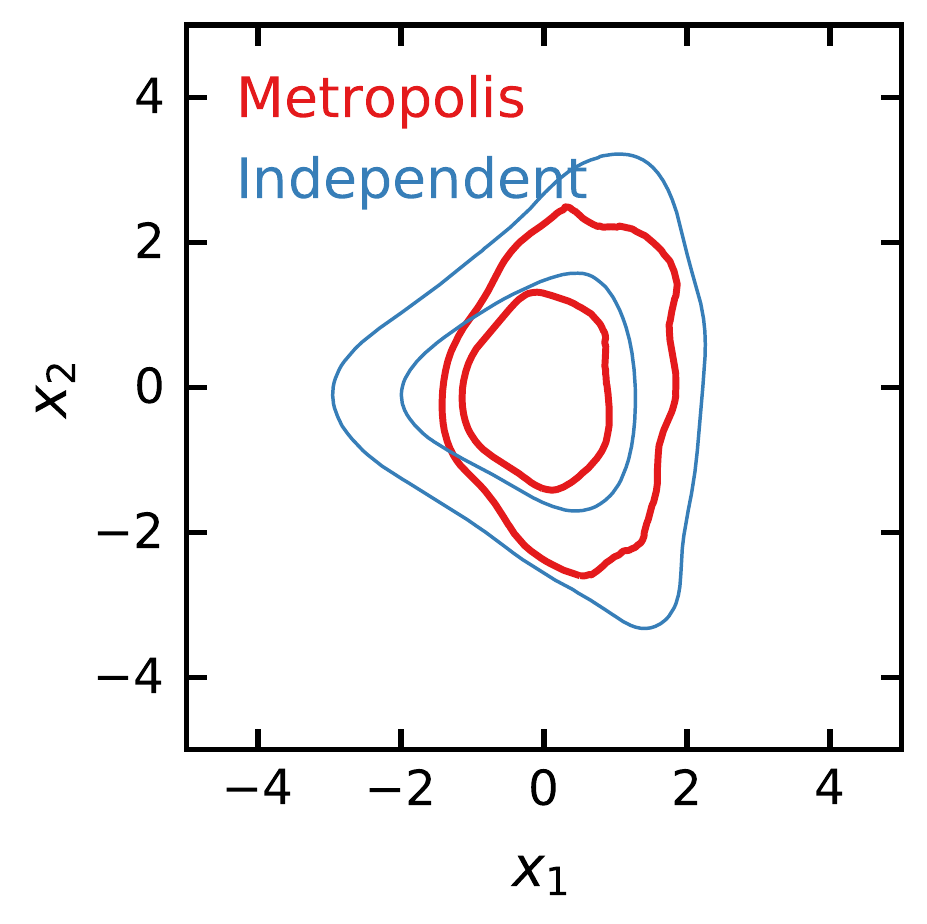}
    \includegraphics[width=.3\textwidth]{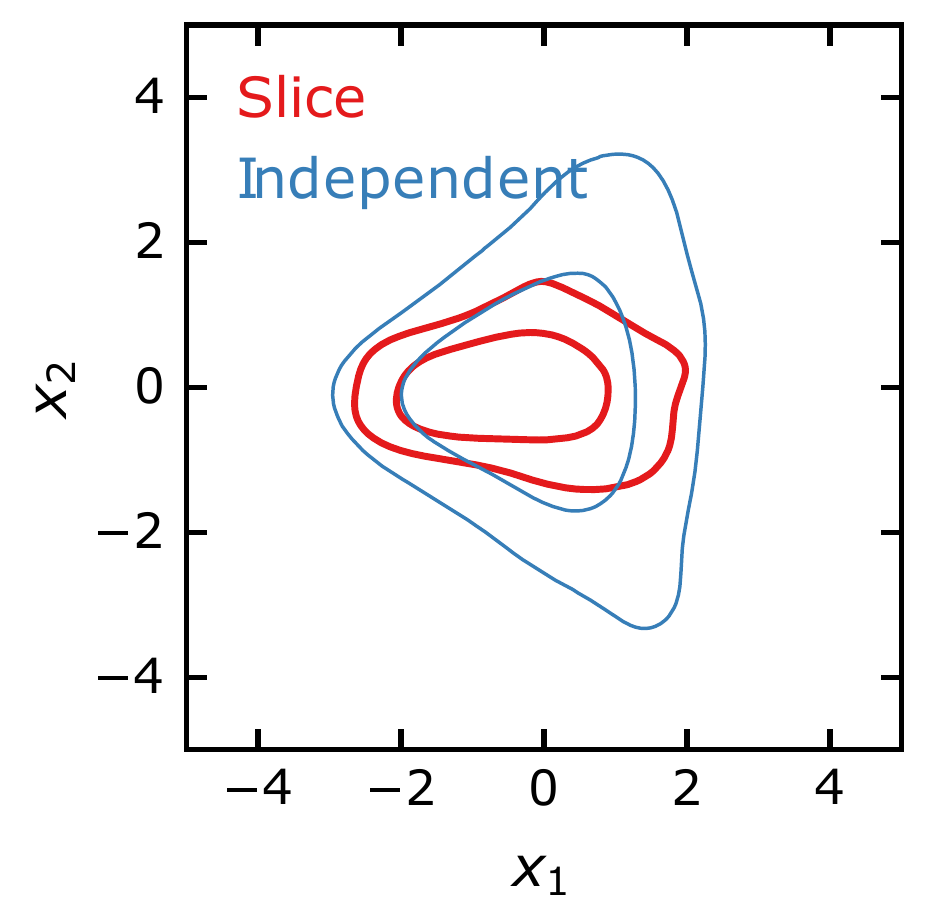}
    \includegraphics[width=.3\textwidth]{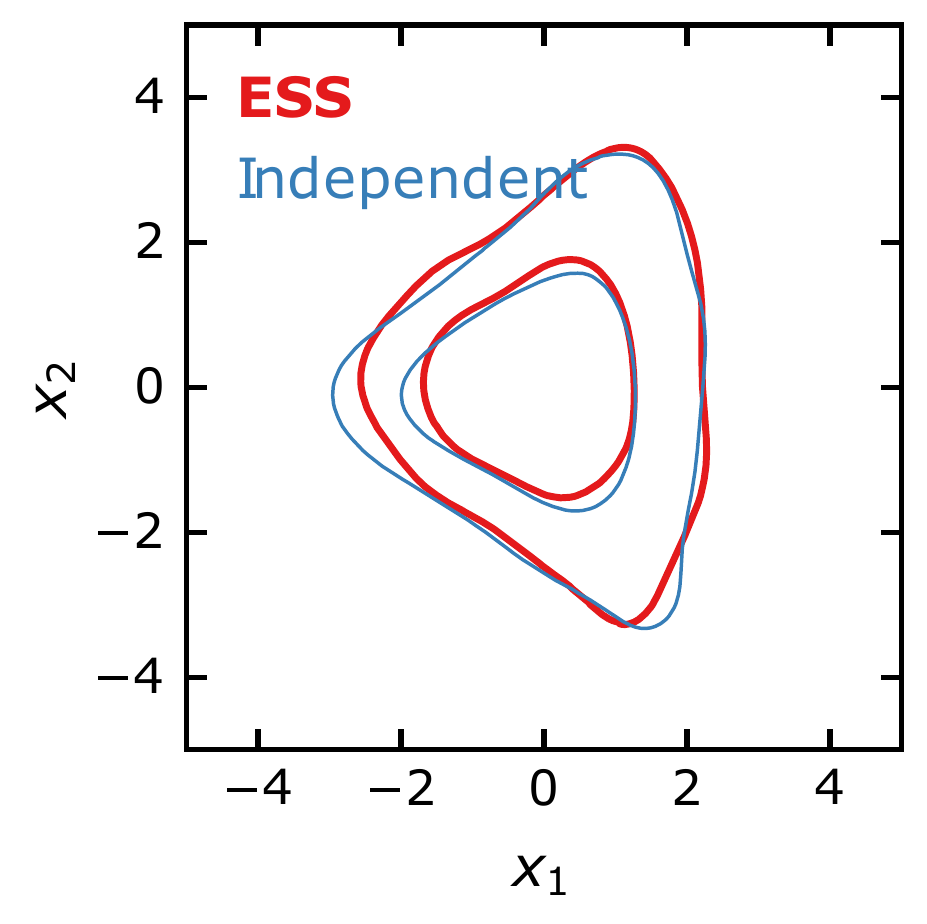}
\caption{The plots compare the 1-sigma and 2-sigma contours generated by the optimised random-walk Metropolis (left), Standard Slice (centre) and Ensemble Slice Sampling (right) methods to those obtained by Independent Sampling (blue) for the correlated funnel distribution. All samplers used the same number of probability density evaluations, $3\times 10^{5}$. Only the first two dimensions are shown here.}
\label{fig:funnel}
\end{figure*}

To assess the mixing rate of the algorithm on this demanding case, we set the maximum number of evaluations of the probability density function to $3\times 10^{5}$. As shown in Figure \ref{fig:funnel}, the Ensemble Slice Sampling is the only algorithm out of the three whose outcome closely resembles the target distribution. The results of Metropolis were incorrect for both, the limited run with $3\times 10^{5}$ iterations and the long run with $10^{7}$ iterations. In particular, the chain produced using the Metropolis method resemble a converged chain but in fact it is biased in favour of positive values of $x_{1}$. The problem arises because of the vanishing low probability of accepting a point with highly negative value of $x_{1}$. This indicates the inability of Metropolis to handle this challenging case. For a more detailed discussion of this problem we direct the reader to Section 8 of \cite{neal2003slice}. In general, the correlated funnel is a clear example of a distribution in which a single Metropolis proposal scale is not sufficient for all the sampled regions of parameter space. The locally adaptive nature of ESS solves this issue.

\subsection{Comparison to other ensemble methods}

So far we have demonstrated Ensemble Slice Sampling's performance in simple, yet challenging, target distributions. The tests performed so far demonstrate ESS's capacity to sample efficiently from highly correlated distributions compared with standard methods such as Metropolis and Slice Sampling. Although the use of Metropolis and Slice Sampling is common, these methods are not considered to be state-of-the-art. For this reason, we will now compare ESS with state-of-the-art gradient-free ensemble MCMC methods.

By far, the two most popular choices\footnote{For instance, in the fields of Astrophysics and Cosmology where most models are not differentiable and gradient methods (e.g. Hamiltonian Monte Carlo or NUTS) are not applicable the default choice is the Affine-Invariant Ensemble Sampler (AIES) \citep{goodman2010ensemble} as implemented in emcee.} of gradient-free ensemble methods are the Affine-Invariant Ensemble Sampling (AIES) \citep{goodman2010ensemble} method and the Differential Evolution Monte Carlo (DEMC) \citep{ter2006markov} algorithm supplemented with a Snooker update \citep{ter2008differential}.

In cases of strongly multimodal target distributions we will also test our method against Sequential Monte Carlo\footnote{As there are many different flavours of SMC, we decided to use the one implemented in \texttt{PyMC3} which utilises importance sampling, simulated annealing and Metropolis sampling.} (SMC) \citep{liu1998sequential, del2006sequential} and Kernel Density Estimate Metropolis (KM) \citep{kombine} which are particle methods specifically designed to handle strongly multimodal densities. \\

\noindent\textbf{Ring distribution:} Although, all three of the compared methods (i.e. ESS, AIES, DEMC) are affine invariant and thus unaffected by linear correlations, they do however differ significantly in the way they handle non-linear correlations. In particular, only Ensemble Slice Sampling (ESS) is locally adaptive because of its stepping-out procedure and therefore able to handle non-linear correlations efficiently.

To illustrate ESS's performance in a case of strong non-linear correlations we will use the 16--dimensional ring distribution defined by:
\begin{equation}
\begin{split}
    \ln \mathcal{L} = & - \Bigg[ \frac{(x_{n}^{2} + x_{1}^{2} - a)^{2}}{b}\Bigg]^{2} \\ & -\sum_{i=1}^{n-1} \Bigg[ \frac{(x_{i}^{2} + x_{i+1}^{2} - a)^{2}}{b}\Bigg]^{2}\, ,
\end{split}
\label{eq:ring}
\end{equation}
where $a=2$, $b=1$ and $n=16$ is the total number of parameters. We also set the number of walkers to be $64$ and run the samplers for $10^{7}$ steps discarding the first half of the chains. Here we followed the heuristics discussed at the beginning of this section and increased the number of walkers from the minimum of $2\times 16$ to $4\times 16$ due to the presence of strong non-linear correlations in order to achieve the optimal acceptance rate for AIES and DEMC. The number of iterations is large enough for all samplers to converge and provide accurate estimates of the autocorrelation time.

The results are shown in Table \ref{tab:table2} and verify that ESS' performance is an order of magnitude better than that of the other methods. \\

\begin{table}[ht!]
    \centering
    \caption{The table shows a comparison of the Affine Invariant Ensemble Sampling (AIES), Differential Evolution Markov Chain (DEMC), and Ensemble Slice Sampling methods in terms of the integrated autocorrelation time (IAT) and the number of effective samples per evaluation of the probability density (efficiency) multiplied by $10^5$. These metrics are formally defined in Appendix \ref{app:ess}. The target distributions are the 16--dimensional ring distribution, the 10--dimensional Gaussian shells distribution and the 13--dimensional hierarchical Gaussian process regression distribution. In all cases the total number of iterations was set to $10^{7}$. It should be noted that in the case of the Gaussian shells the global move was used instead of the differential move.}
    \def\arraystretch{1.1}
    \begin{tabular}{lccc}
        \toprule[0.75pt]
         & AIES   & DEMC   & \textbf{ESS}  \\
        \midrule[0.5pt]
        \multicolumn{4}{l}{Ring distribution} \\
        \midrule[0.5pt]
        IAT          &    49470    &    91128    &   $\mathbf{1675}$   \\
        efficiency   &    2.0    &    1.1    & $\mathbf{12.2}$  \\
        \midrule[0.5pt]
        \multicolumn{4}{l}{Gaussian shells distribution} \\
        \midrule[0.5pt]
        IAT          &    33046    &    2760    &   $\mathbf{89}$   \\
        efficiency   &    3.0    &    36.0    &  $\mathbf{731.0}$ \\
        \midrule[0.5pt]
        \multicolumn{4}{l}{Hierarchical Gaussian process regression} \\
        \midrule[0.5pt]
        IAT          &    55236    &    30990    &   $\mathbf{547}$   \\
        efficiency   &    1.8    &    3.2    &  $\mathbf{38.0}$ \\
        \bottomrule[0.75pt]
        \end{tabular}
    \label{tab:table2}
\end{table}

\noindent\textbf{Gaussian shells distribution:} Another example that demonstrates ESS's performance in cases of non-linear correlations is the Gaussian Shells distribution defined as:
\begin{equation}
    \mathcal{L}(\mathbf{\Theta}) = \text{circ}(\mathbf{\Theta}|\mathbf{c}_{1}, r_{1}, w_{1})+\text{circ}(\mathbf{\Theta}|\mathbf{c}_{2}, r_{2}, w_{2}),
    \label{eq:shells}
\end{equation}
where
\begin{equation}
    \text{circ}(\mathbf{\Theta}|\mathbf{c}, r, w) = \frac{1}{\sqrt{2\pi}w} \exp \Bigg[-\frac{1}{2} \frac{(|\Theta - \mathbf{c}| - r)^{2}}{w^{2}}\Bigg].
    \label{eq:shell}
\end{equation}
We choose the centres, $\mathbf{c}_{1}$ and $\mathbf{c}_{2}$ to be $-3.5$ and $3.5$ in the first dimension respectively and zero in all others. We take the radius to be $r=2.0$ and the width $w=0.1$. In two dimensions, the aforementioned distribution corresponds to two equal-sized Gaussian Shells. In higher dimensions the geometry of the distribution becomes more complicated and the density becomes multimodal.

For our test, we set the number of dimensions to $10$ and the number of walkers to $40$ due to the existence of two modes. Since this target distribution exhibits some mild multimodal behaviour we opt for the global move instead of the default differential move although the latter also performs acceptably in this case. The total number of iterations was set to $10^{7}$ and the first half of the chains was discarded. The results are presented in Table \ref{tab:table2}. ESS's autocorrelation time is $2-3$ orders of magnitude lower than that of the other methods and the efficiency is higher by $1-2$ orders of magnitude respectively. \\

\noindent\textbf{Hierarchical Gaussian process regression:}
To illustrate ESS's performance in a real-world example we will use a modelling problem concerning the concentration of $CO_{2} $ in the atmosphere adapted from Chapter 5 of \cite{rasmussen2003gaussian}. The data consist of monthly measurements of the mean $CO_{2}$ concentration in the atmosphere measured at the \emph{Mauna Loa Observatory} \citep{keeling2004atmospheric} in \emph{Hawaii} since 1958. Our goal is to model the concentration of $CO_{2}$ as a function of time. To this end, we will employ a \emph{hierarchical Gaussian process} model with a composite covariance function designed to take care of the properties of the data. In particular, the covariance function (kernel) is the sum of following four distinct terms:
\begin{equation}
    k_1(r) = \theta_1^2  \exp \left(-\frac{r^2}{2\theta_2} \right)\, ,
    \label{eq:kernel1}
\end{equation}
where $r=x-x'$ that describes the smooth trend of the data,
\begin{equation}
    k_2(r) = \theta_3^2 \exp \left[-\frac{r^2}{2 \theta_4}
                                            -\theta_5\sin^2\left(
                                            \frac{\pi r}{\theta_6}\right)
                                           \right]\, ,
    \label{eq:kernel2}
\end{equation}
that describes the seasonal component,
\begin{equation}
    k_3(r) = \theta_7^2  \left [ 1 + \frac{r^2}{2\theta_8 \theta_9}
                                \right ]^{-\theta_8}\, ,
    \label{eq:kernel3}
\end{equation}
which encodes medium-term irregularities, and finally:
\begin{equation}
    k_4(r) = \theta_{10}^2  \exp \left(-\frac{r^2}{2 \theta_{11}} \right) + \theta_{12}^2\delta_{ij}\, ,
    \label{eq:kernel4}
\end{equation}
that describes the noise. We also fit the mean of the data, having in total 13 parameters to sample.

We sample this target distribution using $36$ walkers for $10^{7}$ iterations and we discard the first half of the chains. The number of walkers that was used corresponds to $1.5$ times the minimum number. We found that this value results in the optimal acceptance rate for AIES and DEMC. For this example we use the differential move of ESS. The results are presented in Table \ref{tab:table2}. The integrated autocorrelation time of ESS is $2$ orders of magnitude lower than that of the other methods and its efficiency is more than an order of magnitude higher. The performance is weakly sensitive to the choice of the number of walkers. \\

\noindent\textbf{Bayesian object detection:} Another real world example with many applications in the field of \emph{astronomy} is \emph{Bayesian object detection}. The following model adapted from \cite{feroz2008multimodal} can be used with a few adjustments to detect astronomical objects in telescope images often hidden in background noise.

We assume that the 2D circular objects present in the image are described by the Gaussian profile:
\begin{equation}
    \mathbf{G}(x,y; \bm{\theta})= A \exp\bigg[-\frac{(x-X)^{2}+(y-Y)^{2}}{2 R^{2}} \bigg]\, ,
    \label{eq:template}
\end{equation}
where $\mathbf{\theta}=(X, Y, A, R)$ are parameters that define the coordinate position, the amplitude and the size of the object, respectively.
Then the data can be described as:
\begin{equation}
    \mathbf{D}=\mathbf{N} + \sum_{i=1}^{n_{\text{Obj}}} \mathbf{G}(\bm{\theta_{i}})\, ,
    \label{eq:data}
\end{equation}
where $n_{\text{Obj}}$ is the number of objects in the image and $\mathbf{N}$ is an additive Gaussian noise term.

\begin{figure}[thb!]
    \centering
    \includegraphics[scale=0.51]{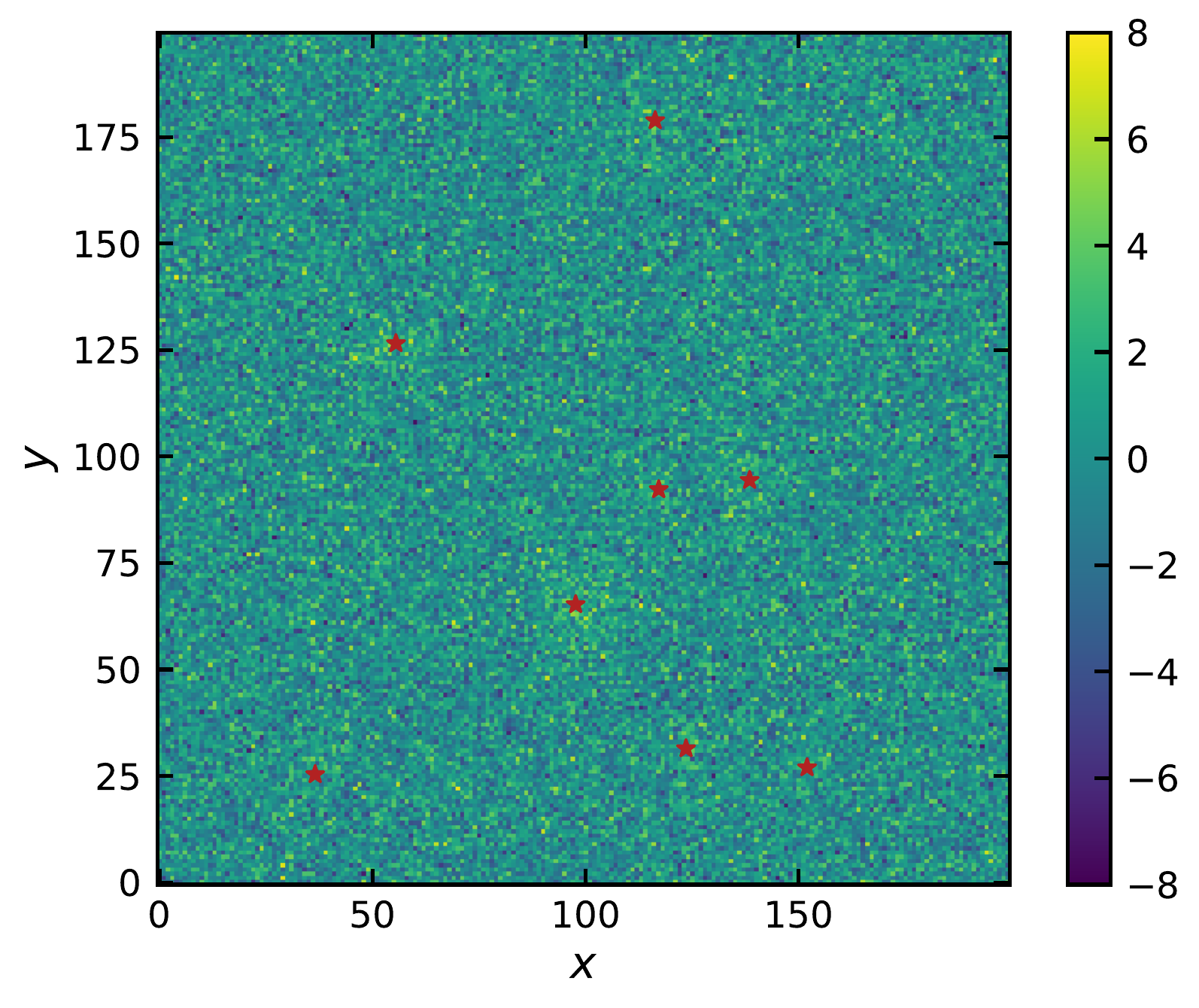}
\caption{The plot shows a simulated image used in the Bayesian object detection exercise. There are $8$ circular objects included here. As the objects are hardly visible due to the background noise their centres are marked with red stars.}
\label{fig:objects}
\end{figure}

Assuming a $200 \times 200$ pixel-wide image, we can create a simulated dataset by sampling the coordinate positions $(X, Y)$ of the objects from $\mathcal{U}(0, 200)$ and their amplitude $A$ and size $R$ from $\mathcal{U}(1, 2)$ and $\mathcal{U}(3, 7)$, respectively. We sample $n_{\text{Obj}} = 8$ objects in total. Finally, we sample the noise $\mathbf{N}$ from $\mathcal{N}(0,4)$. In practice we create a dataset of $100$ such images and one such example is shown in Figure \ref{fig:objects}. Notice that the objects are hardly visible as they are obscured by the background noise, this makes the task of identifying those objects very challenging.

Following the construction of the simulated dataset, the posterior probability density function is defined as:
\begin{equation}
    P(\bm{\theta} | \mathbf{D}) \propto \exp \bigg\{\frac{[\mathbf{G}(\bm{\theta})-\mathbf{D}]^{2}}{2 \sigma^{2}}\bigg\} P(\bm{\theta})\, ,
    \label{eq:obj_post}
\end{equation}
where $\sigma = 2$ is the standard deviation of the $\mathbf{N}$ noise term. The prior $P(\bm{\theta})$ can be decomposed as the product of prior distributions of $X$, $Y$, $A$, and $R$. We used uniform priors for all of these parameters with limits $(0,200)$ for $X$ and $Y$, $(1,2)$ for $A$, and $(2,9)$ for $R$. It is important to mention here that the posterior does not include any prior information about the exact or maximum number of objects in the data. In that sense, the sampler is agnostic about the exact number, positions and characteristics (i.e. amplitude and size) of the objects that it seeks to detect.

We sampled the posterior distribution using $200$ walkers (initialised from the prior distribution) for each image in our dataset (i.e. 100 images in total) using Ensemble Slice Sampling (ESS), Affine Invariant Ensemble Sampling (AIES), and Differential Evolution Markov Chain (DEMC). Although the posterior distribution is multimodal (i.e. $8$ modes) we used the differential move since the number of dimensions is low and there is no reason to use more sophisticated moves like the global move. We used a large enough ensemble of walkers due to the potential presence of multiple modes so that all three samplers are able to resolve them.

We ran each sampler for $10^{4}$ iterations in total and we discarded the first half of the chains. We found that, on average for the 100 images, ESS identifies correctly $7$ out of $8$ objects in the image, whereas AIES and DEMC identify $4$ and $5$, respectively. 

In cases where the objects are well-separated ESS often identifies correctly $8$ out of $8$. Its accuracy falls to $7/8$ in cases where two of the objects are very close to each other or overlap. In those cases ESS identifies the merged object as a single object. \\

\noindent\textbf{Gaussian Mixture:} One strengths of ESS is its ability to sample from strongly multimodal distributions in high dimensions. To demonstrate this, we will utilise a Gaussian Mixture of two components centred at $\mathbf{-0.5}$ and $\mathbf{+0.5}$ with standard deviation of $\mathbf{0.1}$. We also put $1/3$ of the probability mass in one mode and $2/3$ in the other.

We first set this distribution at $10$ dimensions and we sample this using $80$ walkers for $10^{5}$ steps. The distance between the two modes in this case is approximately $32$ standard deviations. We then increase the number of dimensions to $50$ and we sample it using $400$ walkers for $10^{5}$ iterations. In this case, the actual distance between the two modes is approximately $71$ standard deviations. The total number of iterations was set to $10^{7}$ for all methods but the SMC.

This problem consists of two, well separated, modes and thus requires using at least twice the minimum number of walkers (i.e. at least 40 for the 10--dimensional case and 200 for the 50--dimensional one). Although the aforementioned configuration was sufficient for ESS to provide accurate estimates, we opted instead for twice that number (i.e. 80 walkers for the 10--dimensional cases and 400 for the 50--dimensional one) in order to satisfy the requirements of the other samplers, mainly the Kernel Density Estimate Metropolis (KM), but also AIES and DEMC. For the Sequential Monte Carlo (SMC) sampler we used $2000$ and $20000$ independent chains for the low and high dimensional case respectively. The temperature ladder that interpolates between the prior and posterior distribution was chosen adaptively guaranteeing an effective sample size of $90\%$ the physical size of the ensemble. Our implementation of SMC was based on that of \texttt{PyMC3} using an independent Metropolis mutation kernel.

\begin{figure*}[thb!]
    \centering
    \includegraphics[scale=0.55]{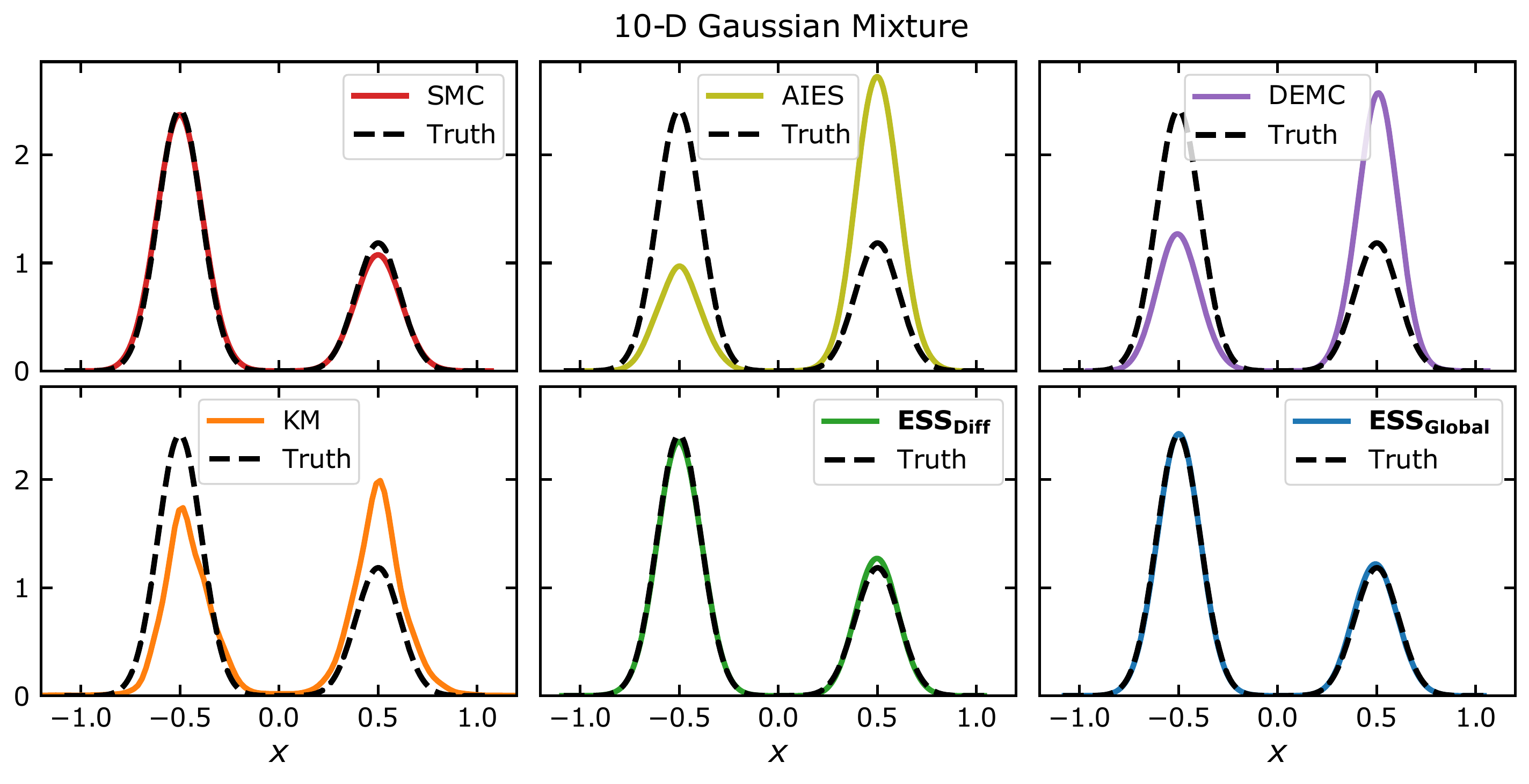}
\caption{The plot compares the results of 6 samplers, namely Sequential Monte Carlo (SMC, red), Affine-Invariant Ensemble Sampling (AIES, yellow), Differential Evolution Markov Chain (DEMC, purple), Kernel Density Estimate Metropolis (KM, orange), Ensemble Slice Sampling using the differential move (ESS, green), and Ensemble Slice Sampling using the global move (ESS, blue). The target distribution is a 10--dimensional Gaussian Mixture. The figure shows the 1D marginal distribution for the first parameter of the 10.}
\label{fig:10dmixture}
\end{figure*}

\begin{figure*}[t!]
    \centering
    \includegraphics[scale=0.55]{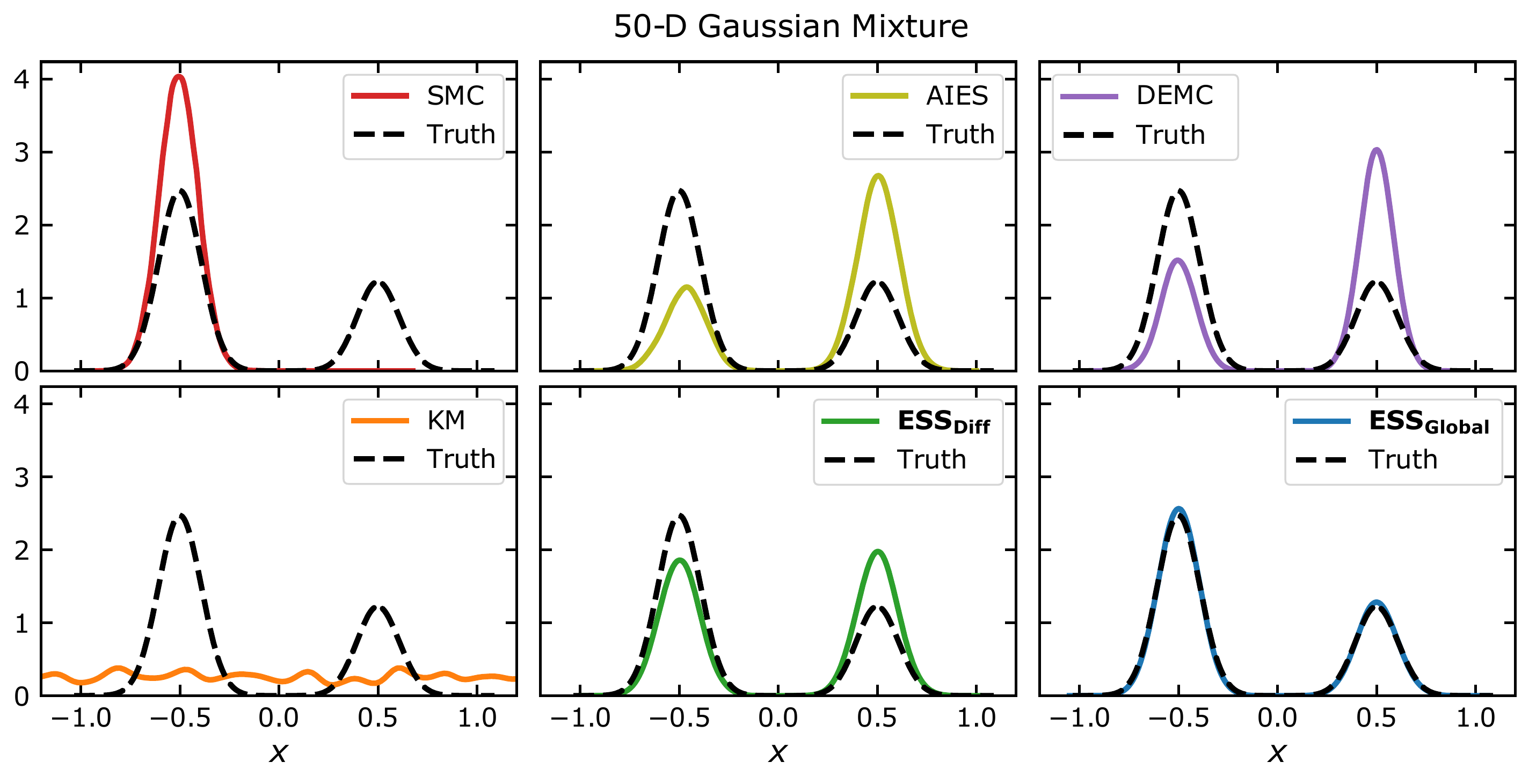}
\caption{The plot compares the results of 6 samplers, namely Sequential Monte Carlo (SMC, red), Affine-Invariant Ensemble Sampling (AIES, yellow), Differential Evolution Markov Chain (DEMC, purple), Kernel Density Estimate Metropolis (KM, orange), Ensemble Slice Sampling using the differential move (ESS, green), and Ensemble Slice Sampling using the global move (ESS, blue). The target distribution is a 50--dimensional Gaussian Mixture.The figure shows the 1D marginal distribution for the first parameter of the 50.}
\label{fig:50dmixture}
\end{figure*}

The results for the 10--dimensional and 50--dimensional cases are plotted in Figures \ref{fig:10dmixture} and \ref{fig:50dmixture}, respectively. In the 10--dimensional case, both ESS (differential and global move) and SMC managed to sample from the target whereas AIES, DEMC and KM failed to do so. In the 50--dimensional case, only the Ensemble Slice Sampling with the global move manages to sample correctly from this challenging target distribution. In practice $\text{ESS}_{G}$ is able to handle similar cases in even higher number of dimensions and with more than $2$ modes.

\subsection{Convergence of the Length Scale \texorpdfstring{$\mu$}{u}}

\begin{figure}[t!]
    \centering
    \includegraphics[scale=0.45]{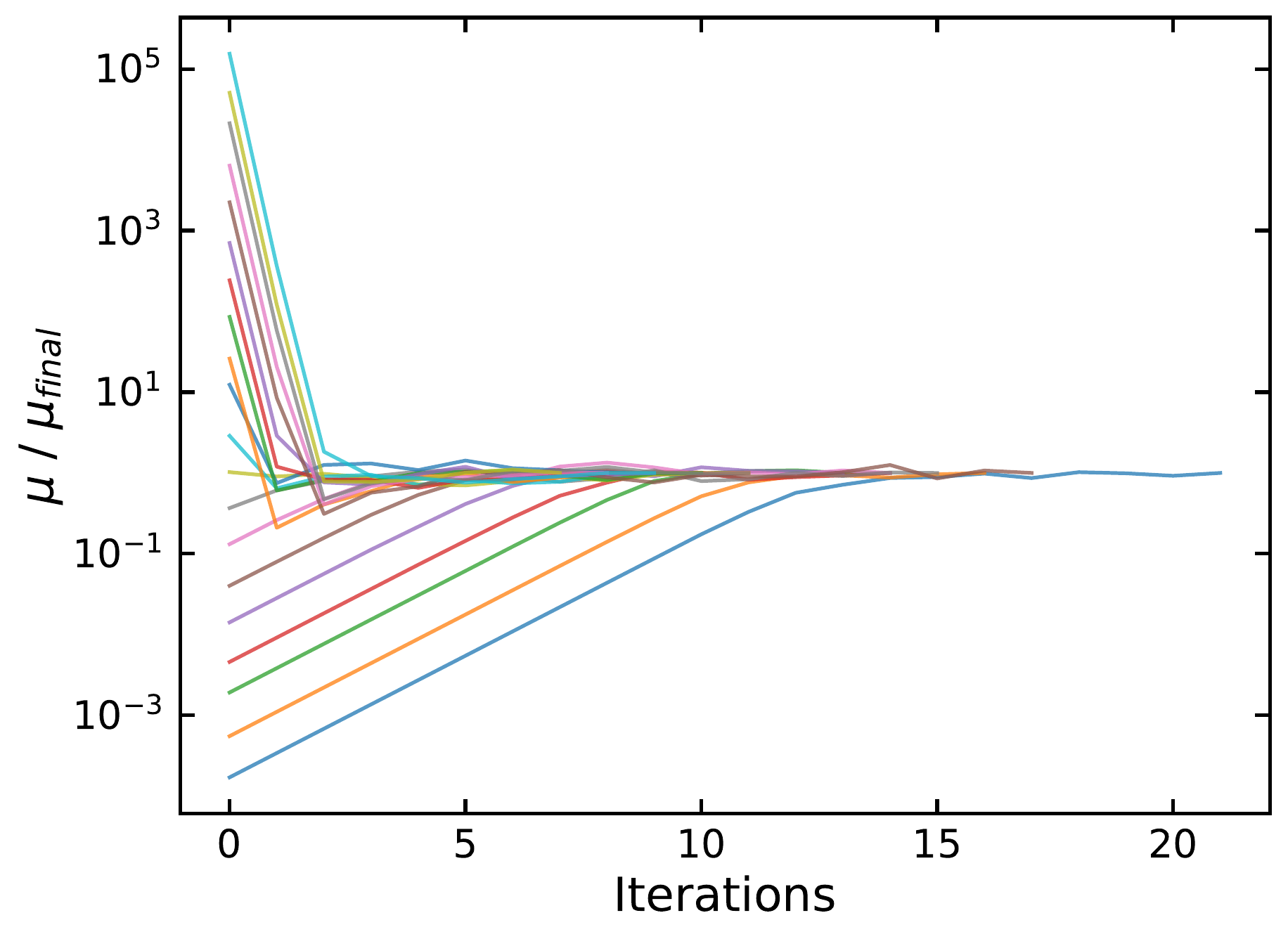}
    \caption{The plot shows the adaptation of the length scale $\mu$ as a function of the number of iterations and starting from a wide range of initial values. Each trace is an independent run and the y-axis shows the value of $\mu$ divided by the final value of $\mu$. The target distribution in this example is a 20--dimensional correlated normal distribution. Starting from larger $\mu$ values leads to significantly faster adaptation.}
\label{fig:scale}
\end{figure}

Figure \ref{fig:scale} plots the convergence of the length scale during the first 20 iterations. The target distribution in this example is a 20--dimensional correlated normal distribution. The length scale $\mu$ was initialised from a wide range of possible values. Adaptation is significantly faster when the initial length scale is larger than the optimal one rather than smaller. Another benefit of using a larger initial estimate would be the reduced number of probability evaluations during the first iterations. This is due to the fact that the shrinking procedure is generally faster than the stepping-out procedure.

\subsection{Parallel Scaling}

\begin{figure}[t!]
    \centering
    \includegraphics[scale=0.45]{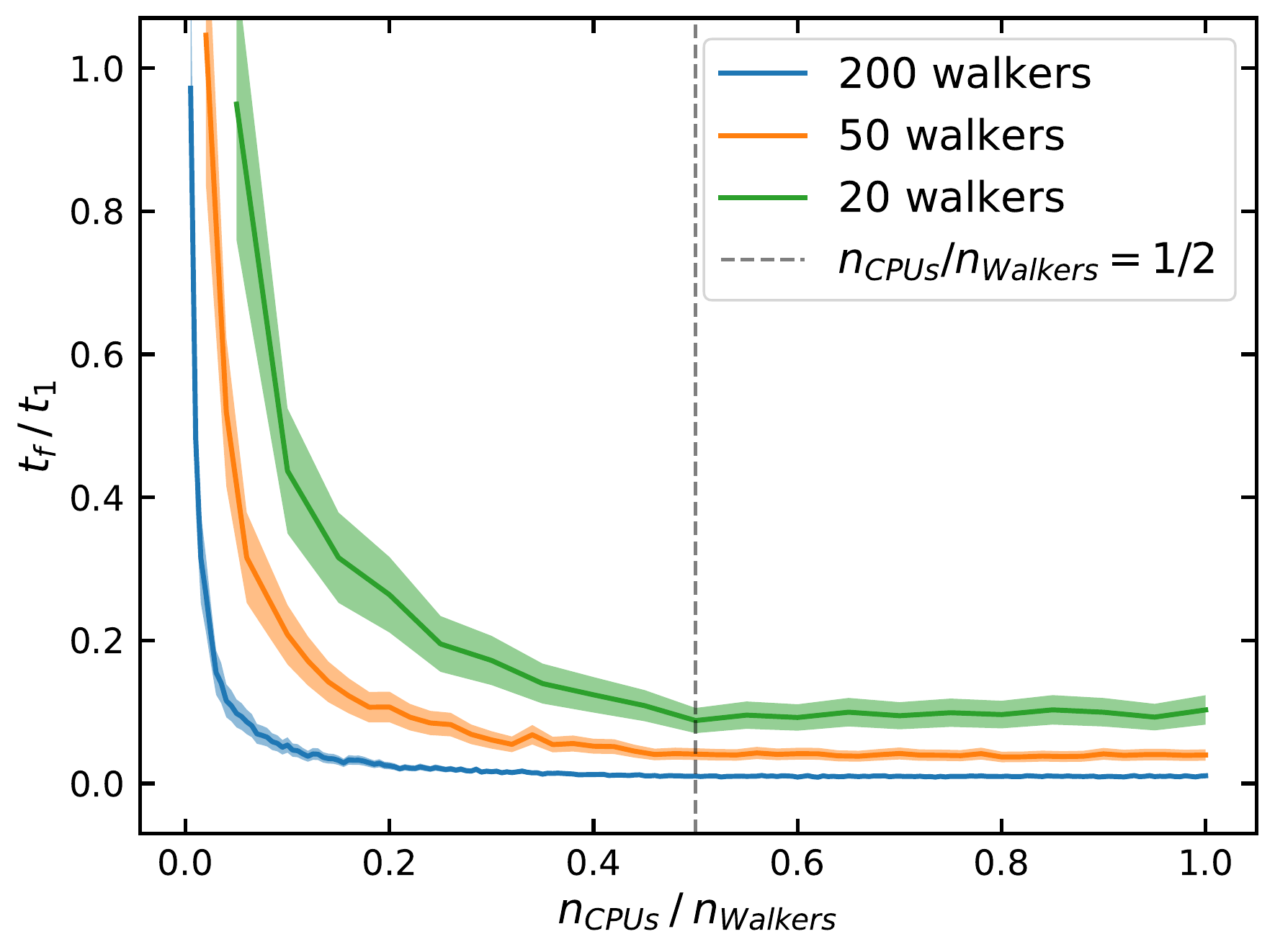}
    \caption{The plot shows the time $t_{f}$ required for ESS to complete a pre-specified number of iterations as a function of the ratio of the number of available CPUs $n_{\rm CPUs}$ to the total number of walkers $n_{\rm Walkers}$. The results are normalised with respect to the single CPU case $t_{1}$. The method scales as $\mathcal{O}(1/n_{\text{CPUs}})$ as long as $n_{\text{CPUs}}\leq n_{\text{Walkers}}/2$ (dashed line). The shaded areas show the $2-\sigma$ intervals.}
\label{fig:parallel}
\end{figure}

By construction, Ensemble Slice Sampling can be used in parallel computing environments by parallelising the ensemble of walkers as discussed in Section \ref{sec:direction}. The maximum number of CPUs used without any of them being idle is equal to the size of complementary ensemble,  $n_{\text{Walkers}}/2$. In order to verify this empirically and investigate the scaling of the method for any number of CPUs, we sampled a 10--dimensional Normal distribution for $10^{5}$ iterations with varying number of walkers. The results are plotted in Figure \ref{fig:parallel}. We sampled the aforementioned distribution multiple times in order to get estimates of the confidence integrals shown in Figure \ref{fig:parallel}. The required time to do the pre-specified number of iterations scales as $\mathcal{O}(1/n_{\text{CPUs}})$ as long as $n_{\text{CPUs}}\leq n_{\text{Walkers}}/2$. This result does not depend on the specific distribution. We can always use all the available CPUs by matching the size of the complementary ensemble (i.e. half the number of walkers) to the number of CPUs.

\section{Discussion}
\label{sec:discussion}

In Section \ref{sec:empirical} we provided a quantitative comparison of the efficiency of Ensemble Slice Sampling compared to other methods. In this Section we will provide some qualitative arguments to informally demonstrate the advantages of Ensemble Slice Sampling over other methods. Furthermore, we will briefly discuss some general aspects of the algorithm and place our work in the context of other related algorithms.

After the brief adaptation period is over and the length scale $\mu$ is fixed, the Ensemble Slice Sampling algorithm performs on average $5$ evaluations of the probability density per walker per iteration, assuming that either the differential or Gaussian move is used. This is in stark contrast with Metropolis-based MCMC methods that perform $1$ evaluation of the probability density per iteration. However, the non-rejection nature of Ensemble Slice Sampling more than compensates for the higher number of evaluations as shown in Section \ref{sec:empirical}, thus yielding a very efficient scheme.

One could think of the number of walkers as the only free hyperparameter of Ensemble Slice Sampling. However, choosing the number of walkers is usually trivial. As we mentioned briefly at the end of Section \ref{sec:ensemble}, there is a minimum limit to that number. In particular, in order for the method to be ergodic, the ensemble should be made of at least $2\times D$ walkers\footnote{The reason that the minimum limit is $2\times D$ instead of $D+1$ has to do with the ensemble splitting procedure that we introduced in order to make the method parallel. Splitting the ensemble into two equal parts means that each walker is updated based on the relative displacements of half the ensemble.}, where $D$ is the number of dimensions of the problem. Assuming that the initial relative displacements of the walkers span the parameter space (i.e. they do not belong to a lower-than-$D$-dimensional space) the resulting algorithm would be ergodic. As shown in Section \ref{sec:empirical}, using a value close to the minimum number of walkers, meaning twice the number of parameters, is generally a good choice. Furthermore, we suggest to increase the number of walkers by a multiplicative factor equal to the number of well separated modes (e.g. four times the number of dimensions in a bimodal density). Other cases in which increasing the number of walkers can improve the sampling efficiency include target distributions with strong non-linear correlations between their parameters.

Regarding the initial positions of the walkers, we found that we can reduce the length of the burn-in phase by initialising the walkers from a tight sphere (i.e. Normal distribution with a very small variance) close to the \emph{Maximum a Posteriori} (MAP) estimate. In high dimensional problems, the MAP estimate will not reside in the typical set and the burn-in phase might be longer. We found that the tight sphere initialisation is still an efficient strategy compared to a more dispersed initialisation \citep{foreman2013emcee}. Other approaches include initialising the walkers by sampling from the prior distribution or the \emph{Laplace approximation} of the posterior distribution. In multimodal cases, a prior initialisation is usually a better choice. A brief simulated annealing phase can also be very efficient, particularly in cases with many well separated modes.

Recent work on the No U-Turn Sampler \citep{hoffman2014no} has attempted to reduce the hand-tuning requirements of Hamiltonian Monte Carlo \citep{betancourt2017conceptual} using the dual averaging scheme of \citet{nesterov2009primal}. In order to achieve a similar result, we employed the much simpler stochastic approximation method of \citet{robbins1951stochastic} to tune the initial length scale $\mu$. The Affine Invariant Ensemble Sampler \citep{goodman2010ensemble} and the Differential Evolution MCMC \citep{ter2006markov} use an ensemble of walkers to perform Metropolis updates. Our method differs by using the information from the ensemble to perform Slice Sampling updates. So why does ESS perform better, as demonstrated, compared to those other methods? The answer lies in the locally adaptive and non-rejection nature of the algorithm (i.e. stepping out and shrinking) that enables both efficient exploration of non-linear correlations and large steps in parameter space (e.g. using the global move)\footnote{Indeed, large steps like the ones in the 50--dimensional Gaussian Mixture example would not have been possible without the non-rejection aspect of the method as most attempts to jump to the other mode would have missed it using Metropolis updates.}.

For all numerical benchmarks in this paper we used the publicly available, open source \texttt{Python} implementation of Ensemble Slice Sampling called  \texttt{zeus}\footnote{The code is available at \url{https://github.com/minaskar/zeus}.}~\citep{karamanis2021zeus}.

\section{Conclusion}
\label{sec:conclusion}

We have presented Ensemble Slice Sampling (ESS), an extension of Standard Slice Sampling that eliminates the latter's dependence on the initial value of the length scale hyperparameter and augments its capacity to sample efficiently and in parallel from highly correlated and strongly multimodal distributions. 

In this paper we have compared Ensemble Slice Sampling with the optimally-tuned Metropolis and Standard Slice Sampling algorithms. We found that, due to its affine invariance, Ensemble Slice Sampling generally converges faster to the target distribution and generates chains of significantly lower autocorrelation. In particular, we found that in the case of AR(1), Ensemble Slice Sampling generates an order of magnitude more independent samples per evaluation of the probability density than Metropolis and Standard Slice Sampling. Similarly, in the case of the correlated funnel distribution, Ensemble Slice Sampling outperforms Standard Slice Sampling by an order of magnitude in terms of efficiency. Furthermore, in this case, Metropolis-based proposals fail to converge at all, demonstrating that a single Metropolis proposal scale is often not sufficient.

When compared to state-of-the-art ensemble methods (i.e. AIES, DEMC) Ensemble Slice Sampling outperforms them by $1-2$ orders of magnitude in terms of efficiency for target distributions with non-linear correlations (e.g. the Ring and Gaussian shells distributions). In the real world example of hierarchical Gaussian process regression, ESS's efficiency is again superior by $1-2$ orders of magnitude. Furthermore, in the Bayesian object detection example ESS achieved higher accuracy compared to AIES and DEMC. Finally, in the strongly multimodal case of the Gaussian Mixture, ESS outperformed all other methods (i.e. SMC, AIES, DEMC, KM) and was the only sampler able to produce reliable results in $50$ dimensions.

The consistent high efficiency of the algorithm across a broad range of different problems along with its parallel, black-box and gradient-free nature, renders Ensemble Slice Sampling ideal for use in scientific fields such as physics, astrophysics and cosmology, which are dominated by a wide range of computationally expensive and almost always non-differentiable models. The method is flexible and can be extended further using for example tempered transitions~\citep{iba2001extended} or subspace sampling~\citep{vrugt2009accelerating}.


\begin{acknowledgements}
The authors thank Iain Murray and John Peacock for providing constructive comments on an early draft. The authors would also like to extend their gratitude to the anonymous reviewer and editor for providing comments that helped improve the quality of the manuscript. FB is a Royal Society University Research Fellow. FB is supported by the European Research Council (ERC) under the European Union’s Horizon 2020 research and innovation programme (Grant agreement No. 853291).
\end{acknowledgements}



\appendix
\section{Estimating the Effective Sample Size}
\label{app:ess}



Assuming that the computational bottleneck of a MCMC analysis is the evaluation of the probability density function, which is usually a valid assumption in scientific applications, the \emph{efficiency} can be formally defined as the ratio of the \textit{Effective Sample Size} $N_{\rm Eff}$ to the total number of probability evaluations for a given chain. 

The $N_{\rm Eff}$ quantifies the number of effectively independent samples of a chain, and it is defined as
\begin{equation}
    \label{eq:ESS}
    N_{\rm Eff} = \frac{n}{\text{IAT}}\, ,
\end{equation}
where $n$ is the actual number of samples in the chain, and IAT is the \textit{integrated autocorrelation time}. The latter describes the number of steps that the sampler needs to do in order to forget where it started and it is defined as
\begin{equation}
    \label{eq:act}
    \text{IAT} = 1 + 2\sum_{k=1}^{\infty}\rho(k)\, ,
\end{equation}
where $\rho(k)$ is the \textit{normalised autocorrelation function} at lag $k$. In practise, we truncate the above summation in order to remove noise from the estimate \citep{sokal1997monte}.

Given a chain $X(k)$ with $k=1,2,...,n$ the normalised autocorrelation function $\Hat{\rho}(k)$ at lag $k$ is estimated as 
\begin{equation}
    \Hat{\rho}(k) = \frac{\Hat{c}(k)}{\Hat{c}(0)}\, ,
\end{equation}
where
\begin{equation}
    \Hat{c}(k) = \frac{1}{n-k}\sum_{m=1}^{n-k}\big[ X(k+m)- \Bar{X} \big]\big[  X(m)-\Bar{X} \big]\, ,
\end{equation}
and $\Bar{X}$ is the mean of the samples.

In the case of ensemble methods, the IAT of an ensemble of chains is computed by first concatenating the chain from each walker into a single long chain. We found this estimator has lower variance than the \cite{goodman2010ensemble} estimator and the \cite{noauthor_autocorrelation_nodate} estimator. 
%



%
%

\bibliographystyle{spbasic}      
\bibliography{sample.bib}   

%
%

\end{document}